\documentclass[12pt]{article}

\usepackage[T1]{fontenc}
\usepackage{lmodern}

\usepackage[margin=1in]{geometry}

\usepackage[authoryear,round]{natbib}
\let\cite\citep

\usepackage[toc,page]{appendix}

\usepackage{amsmath,amssymb}
\usepackage{bm}
\usepackage{graphicx}
\usepackage{booktabs}
\usepackage{tabularx}
\usepackage{array}
\usepackage{hyperref}
\usepackage{url}
\usepackage{subcaption}
\usepackage{caption}
\usepackage{tikz}
\usetikzlibrary{shapes.geometric, arrows.meta, positioning, shadows, backgrounds, calc, decorations.pathreplacing}
\usepackage{colortbl}
\usepackage{multirow}

\definecolor{headerblue}{HTML}{1F4788}
\definecolor{lightblue}{HTML}{E8F4FD}
\definecolor{lightgray}{HTML}{F5F7FA}

\graphicspath{{figures/}}

\newcommand{\cC}{\mathcal{C}}
\newcommand{\nullnotation}{\mathsf{N/A}}

\newcommand{\estimator}{\hat{y}}
\newcommand{\estimatorOrig}{\estimator^{\mathrm{original}}}
\newcommand{\estimatorNew}{\estimator^{\mathrm{new}}}
\newcommand{\estimatorFinal}{\estimator^{\mathrm{final}}}

\begin{document}

\title{Scalable Stewardship of an LLM-Assisted Clinical Benchmark with Physician Oversight}

\author{%
Junze (Tony) Ye$^{1}$,\quad
Daniel Tawfik$^{2}$,\quad
Alex J.\ Goodell$^{3}$,\quad
Nikhil V.\ Kotha$^{4}$,\\[4pt]
Mark K.\ Buyyounouski$^{4}$,\quad
Mohsen Bayati$^{1,4,5}$%
}

\date{%
\vspace{6pt}\small
$^{1}$Department of Operations, Information and Technology, Stanford Graduate School of Business, Stanford, CA, USA\\
$^{2}$Department of Pediatrics, Division of Critical Care Medicine, Stanford University School of Medicine, Stanford, CA, USA\\
$^{3}$Department of Anesthesiology, Perioperative and Pain Medicine, Stanford University School of Medicine, Stanford, CA, USA\\
$^{4}$Department of Radiation Oncology, Stanford University School of Medicine, Stanford, CA, USA\\
$^{5}$Department of Electrical Engineering, Stanford University School of Engineering, Stanford, CA, USA%
}

\maketitle

\begin{abstract}
Reference labels for machine-learning benchmarks are increasingly synthesized with LLM assistance, but their reliability remains underexamined. We audit MedCalc-Bench, a clinical benchmark for medical score computation whose labels were partly derived with LLM assistance, and develop a scalable physician-in-the-loop stewardship pipeline to reassess them. At least 27\% of test labels are likely erroneous or incomputable. On a 50-instance subset validated by physicians, our recomputed labels agree with physician ground truth 74\% of the time (95\% CI, 60--84\%) versus 20\% for the originals (95\% CI, 11--33\%). Using original labels to evaluate frontier LLMs underestimates accuracy by 16--23 percentage points. In a controlled reinforcement-learning experiment, a model trained on recomputed labels outperforms one trained on originals by 13.5 percentage points (95\% CI, 10.6--16.6\%) on physician-labeled instances, and this advantage extends to related medical tasks. LLM-assisted benchmarks can propagate systematic errors into both evaluation and post-training unless actively stewarded.
\end{abstract}

\section{Introduction}\label{sec:intro}

Benchmarks have become the primary infrastructure for measuring AI capabilities. The standard recipe pairs a set of tasks with reference answers and a verification procedure, producing a score that summarizes model performance on a given capability \cite{medhelm}. As large language models (LLMs) are deployed in higher-stakes settings, the demand for domain-specific benchmarks has grown accordingly: in medicine alone, recent work has introduced benchmarks for clinical question answering, diagnostic reasoning, and risk-score computation \cite{medqa, medcalc, goodell2025}. These benchmarks shape not only how models are evaluated but increasingly how they are trained, as benchmark labels are reused as reward signals for reinforcement learning (RL) and other post-training methods \cite{ouyang2022, ahmadian2024}.

Building benchmarks at sufficient scale is itself a major challenge. Statistically meaningful evaluation requires large sample sizes, yet producing reference labels in specialized domains demands expensive expert labor: in medicine, each label may require a physician to review the patient's chart and exert clinical judgment to arrive at an answer. Fully physician-labeled datasets therefore tend to be small \cite{goodell2025, roeschl2025}. This tension between scale and quality has made it natural, even necessary, to enlist LLMs in the benchmark-creation process itself. The result is a circularity: AI systems help build the very evaluation infrastructure used to judge AI systems. Label errors in benchmarks are not hypothetical: prior work has shown that commonly used ML datasets harbor pervasive mislabeling that can destabilize model rankings \cite{northcutt2021labelerrors}, and recent studies have used LLM ensembles to detect and correct such errors in NLP benchmarks \cite{nahum2024labelerrors}. However, in high-stakes domains such as medicine, where expert labeling costs are high and the consequences of benchmark error are particularly severe, systematic evidence remains scarce.

We provide such evidence in one of the high-stakes domains. Medical risk scores such as LACE, CURB-65, and CHA\textsubscript{2}DS\textsubscript{2}-VASc condense patient data into numeric values that inform triage, admission, and preventive therapy \cite{lace,lim2003curb65,lip2010cha2ds2vasc}. Automating their computation from clinical narratives is a natural target for LLMs \cite{tierney2024ambient,afshar2025pragmatic,medic-mohsen,korom2025penda,chung2025}, and MedCalc-Bench \cite{medcalc} is the most widely used public benchmark for this task. Its labeling pipeline used GPT-4, the leading model at the time of the benchmark's construction, for feature extraction and scripted aggregation to produce 11{,}100 labeled instances across 55 calculators. The benchmark has been adopted broadly in academic research \cite{medhelm, riskagent, ehrmind, med-u1, score-to-steps, oh2025} and was one of two medical benchmarks evaluated on by Anthropic in its recent ``Claude for Healthcare'' launch \cite{anthropic_press, anthropic_youtube}. Despite this reach, the clinical accuracy of its LLM-generated labels has not been systematically assessed.

In this paper, we develop a scalable stewardship protocol and apply it to MedCalc-Bench. At its core, multiple tool-augmented LLM agents independently recompute each score from the patient narrative alone, without access to the original label, and abstain when the clinical context is insufficient or mismatched to the question. A consensus mechanism across agents produces high-confidence recomputed labels. Automated triage then identifies instances where the recomputed and original labels diverge most, concentrating scarce physician time on the highest-impact disagreements; physicians independently recompute scores for these contentious cases, providing ground-truth adjudication.

We find that a substantial fraction of MedCalc-Bench test labels are likely erroneous or incomputable. We examine the downstream consequences of these errors on two fronts: first, by re-evaluating today's frontier LLMs against our corrected labels and comparing with officially reported results, including those from Anthropic's ``Claude for Healthcare'' launch \cite{anthropic_press}; and second, through a controlled RL experiment in which the only variable is whether training rewards are derived from the original or corrected labels. Both analyses demonstrate that label quality materially affects conclusions about model capabilities. MedCalc-Bench was a significant undertaking that laid the foundation for studying LLM performance on medical score computation; we view our work as building on that foundation rather than critiquing it, offering a transparent, repeatable stewardship process and arguing that benchmarks in safety-critical domains are best maintained as living documents rather than treated as static artifacts.
\section{Methods}

\subsection{Benchmark setting and task definition}
MedCalc-Bench is a public benchmark designed to evaluate whether language models can compute common medical scores from free-text narratives, without requiring external calculator use \cite{medcalc} (a worked example using the LACE score is provided in the Results section and Table~\ref{tab:formalism-lace-mapping} in Supplementary Appendix~\S\ref{sec:medical-score}). Each benchmark instance presents a model with a clinical context, typically several paragraphs of a de-identified patient narrative, and a question requesting a specific medical score (e.g., ``What is the patient's HAS-BLED score?''). The model must return either a numeric value, a datetime, or, in our extended formulation, an explicit abstention (``$\nullnotation$'') when the context lacks necessary information or is clinically mismatched to the question. The benchmark contains 11{,}000 context-question pairs (10{,}053 training; 1{,}047 test) covering a set of 55 different score questions. This task mirrors what clinicians do: read documentation, identify the criteria a particular score requires, and apply the scoring rule. The benchmark therefore tests whether a model can perform both subtasks, criterion extraction from unstructured text and rule-based aggregation, correctly and in sequence.

\paragraph{Dataset versioning.}\label{methods:dataset-versioning} Our analysis focuses on the peer-reviewed release of MedCalc-Bench (v1.0), which serves as the standard baseline for current academic and industry evaluations (e.g., \cite{medhelm}, \cite{anthropic_press}; see also the last paragraph of Appendix \S\ref{sec:current-standard}). We note that while the authors began releasing updates (v1.1, v1.2) during the final preparation of this manuscript, our auditing and reinforcement learning experiments were completed prior to these releases. We therefore utilize v1.0 to ensure: (1) comparability with the existing body of concurrent work, and (2) that our stewardship protocol is tested against the distribution of errors present in the widely circulated version of the benchmark. A detailed analysis of the performance discrepancies between dataset versions is provided in Supplementary Appendices \ref{app:concur-work} and \ref{app:claude}.

\paragraph{Label generation and correctness criteria.} In MedCalc-Bench's publication \cite{medcalc}, reference labels were produced by a three-stage procedure: GPT-3.5 matched patient contexts to appropriate calculator questions, GPT-4 extracted clinical features from each narrative (e.g., patient demographics, symptoms, laboratory values), and custom Python scripts aggregated those features according to each calculator's formula or decision logic. While GPT-4 was among the most capable models available when MedCalc-Bench was created and LLMs have improved substantially since then, even today's most capable models in agentic settings still hallucinate and make errors on clinical tasks, underscoring the need for systematic verification of LLM-assisted labels regardless of the model used to produce them. At test time, a model prediction is scored correct if it falls within $\pm$5\% of the reference label for numeric outputs or matches exactly for datetimes. We extend this scheme to treat abstention as correct when our recomputed reference label is itself ``unknown.''

\paragraph{Reference label as an estimator.} We define \emph{physician-aligned ground truth}, $y^\star$, as the answer a conscientious physician would provide for a given note-question instance $(C,q)$. The corresponding benchmark reference label, which we denote by $\hat{y}$, is an \emph{estimate} of this (latent) ground truth; $\estimatorOrig$ is the reference label originally given by MedCalc-Bench, while $\estimatorNew$ is the label our agentic workflow independently recomputes in \S\ref{sec:phase2_relabel}. This two-stage structure, feature extraction followed by rule-based aggregation, is formalized in Supplementary Appendix~\S\ref{subsec:math}. Estimation error can enter at either stage, whether through inappropriate context-calculator pairing, hallucinated or misread feature values, or aggregation scripts that diverge from published scoring rules. Task-level ambiguity, such as missing timepoints or guideline version differences, introduces additional uncertainty. Recognizing these labels as fallible estimates, rather than one-off oracles, motivates the stewardship process described below.

\paragraph{Notation for label indexing.}
Throughout this paper, $y$ denotes a \emph{single}  label for an instance $(C,q)$, regardless of whether it contains a subscripted index. When discussing multiple instances, we use subscripts ($y^\star_i$, $\estimatorOrig_i$, or $\estimatorNew_i$) to disambiguate the label associated with instance $i$. When the instance is clear from context (e.g., there is only one $(C,q)$), we drop the ``$_i$'' subscript and write $y^\star$, $\estimatorOrig$, or $\estimatorNew$ for readability.

\subsection{Stewardship process overview}
The central constraint in validating benchmark reference labels is not machine computation but physician experts' attention. GPU hours are elastic; physician hours are not. Our process is therefore designed around a principle of \emph{phased resource commitment}: each phase produces evidence that justifies, or gates, incremental investment, concentrating scarce expert time on instances most likely to benefit from it.

\begin{figure}[htbp]
    \centering
    \includegraphics[width=1.0\linewidth]{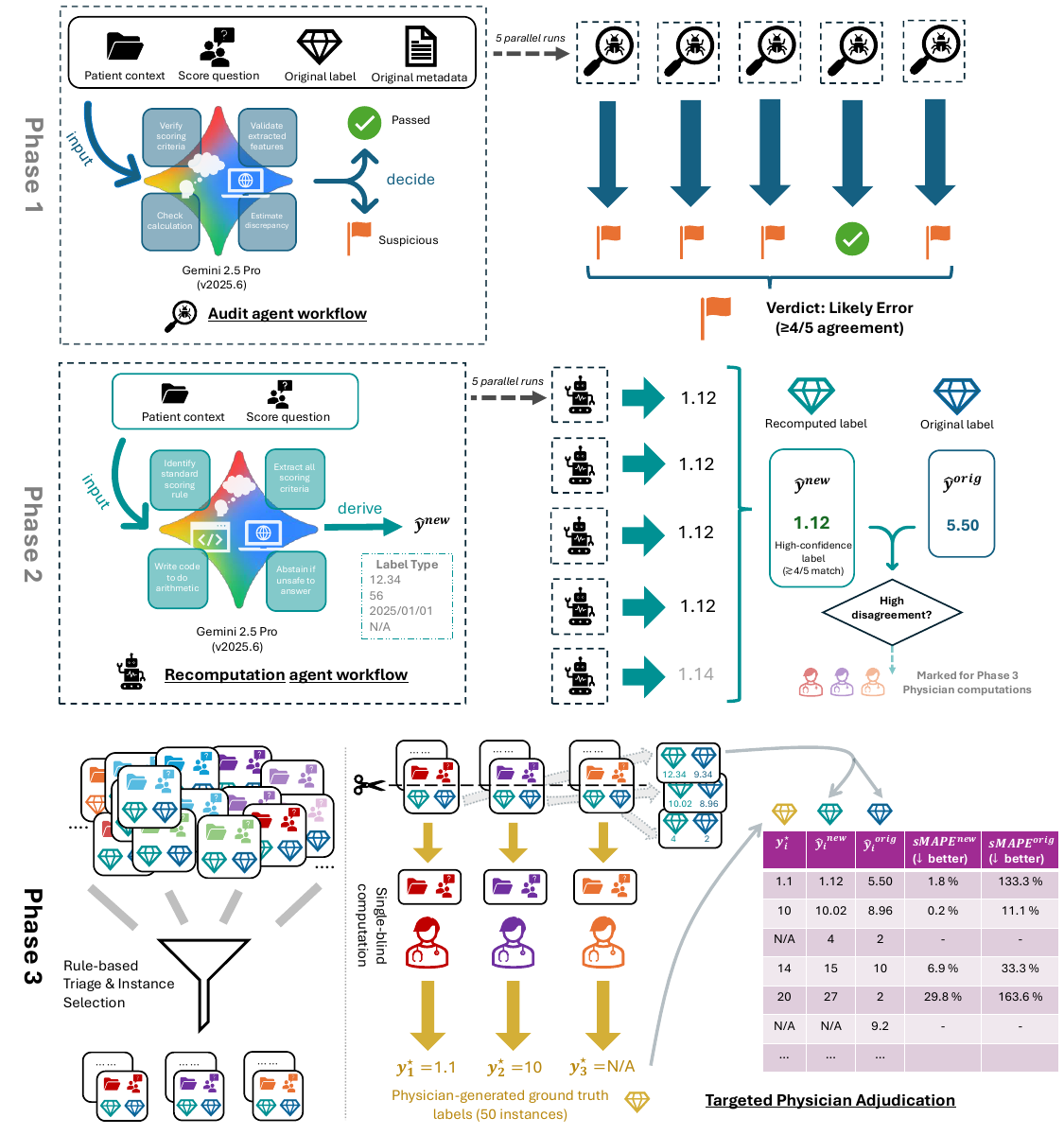}
    \caption{Three-phase stewardship workflow. \textbf{Phase 1} (audit): an LLM verifier reviews each MedCalc-Bench label against its derivation metadata and flags clinically suspect cases. \textbf{Phase 2} (recomputation): a separate agent recomputes the label from the patient narrative and question alone, producing $\estimatorNew$; the original label is hidden to prevent anchoring. Phases 1 and 2 each run five LLM agents independently in parallel and aggregate their outputs by supermajority voting to reduce stochastic noise. \textbf{Phase 3} (adjudication): three physician authors (DT, AG, NK) blind-recompute the 50 instances where $\estimatorOrig$ and $\estimatorNew$ disagree most, producing physician ground truth $y^\star$. sMAPE measures the relative gap between each label set and $y^\star$ (Eq.~\eqref{eq:smape} in \S\ref{sec:phase3-physician}).}
    \label{fig:process-summary}
\end{figure}

The process proceeds in three phases (Figure~\ref{fig:process-summary}), in which the first two serve as an automated instances triage. \textbf{Phase 1 (automated audit)} applies a tool-augmented LLM verifier to existing labels, flagging instances where the original derivation of $\estimatorOrig$ appears clinically unsound. This phase is inexpensive to run and does not commit to replacement labels; its purpose is to provide an early signal of the overall error rate, justifying further investigation. \textbf{Phase 2 (independent recomputation)} deploys a separate LLM pipeline that sees only the patient context and question, not the original label or its provenance, and recomputes a label $\estimatorNew$ from first principles. Disagreement between the recomputed label and the original, beyond a clinically meaningful threshold, identifies cases where at least one estimate is wrong. Finally, \textbf{Phase 3 (physician adjudication)} directs clinician review to the subset of instances where Phase 2 recomputation diverges most substantially from original labels, maximizing information gained per expert hour.

\subsection{Phase 1: Tool-augmented automated audit of original labels}
\label{sec:phase1_audit}

\paragraph{Objective.}
Phase~1 is a low-cost automated audit of MedCalc-Bench test labels. Its purpose is to provide an early signal of whether the benchmark’s original label-derivation procedure produces clinically meaningful inconsistencies, and hence whether it is worth investing in the more resource-intensive Phases~2--3. The fraction of $\{\estimatorOrig_i\}_{i=1}^{1047}$ identified in this phase as problematic also serves as a sensitivity check of the problematic fraction implied by independent label recomputation in Phase~2, which has a different agentic workflow setup.

\paragraph{Audit inputs and label-derivation metadata.}
We built an LLM auditor agent for the MedCalc-Bench \emph{test} split (1{,}047 instances). For each instance, the auditor received: (1) the patient context $C$ (a de-identified clinical narrative), (2) the calculator question $q$, (3) the original benchmark label $\estimatorOrig$, and (4) original metadata describing \emph{how MedCalc-Bench derived} $\estimatorOrig$. (4) includes the structured clinical features extracted from $C$ by the original pipeline (e.g., laboratory values, vitals, demographics) and a description of the calculator-specific aggregation logic used to map extracted features to $\estimatorOrig$. Supplying these artifacts enables the auditor to evaluate whether $\estimatorOrig$ is consistent with both the clinical narrative and a clinically sound extraction-and-aggregation procedure for score question $q$.

\paragraph{Auditing agent and tool access.}\label{methods:model-choice}
We implemented the auditor as a reasoning language model (Gemini 2.5 Pro, June 2025 version) configured as a verifier with web-search access. We selected Gemini 2.5 Pro because at the time Phases 1 and 2 of this study were conducted (June--October 2025), it was among the highest-ranked models on a wide range of public reasoning and knowledge benchmarks, including those covering scientific and clinical domains. The same model is reused as the Phase 2 recomputation agent (\S\ref{sec:phase2_relabel}) to keep the labeling pipeline internally consistent. We separately evaluate today's frontier LLMs (released after our auditing experiments were completed) against the resulting curated labels in \S\ref{results:eval-frontier}. During each audit, the agent could retrieve calculator definitions, rule tables, and guideline statements as needed, improving factual grounding beyond parametric knowledge.

\paragraph{Audit procedure and conservative flagging.}
Each Phase 1 audit run was designed to verify the benchmark’s existing label and its derivation artifacts, not to generate a new blinded label.
Concretely, the auditor (i) identified the appropriate scoring rule for $q$, using web search when needed; (ii) checked whether the required clinical criteria are present and interpretable in $C$; (iii) validated that the structured feature values reported in the benchmark metadata agree with the evidence stated in $C$; and (iv) performed a sanity-check calculation under the identified rule.
The agent was instructed to be conservative and to flag \emph{Suspicious} only when it could articulate a clinically meaningful inconsistency.

\paragraph{Reliability controls via repeated sampling and supermajority voting.}
We use a sampling temperature of 1 for generating outputs, following Google's official recommendation \cite{google_ai_param}. Because LLM outputs are stochastic, we ran the auditing agent five times independently for each $(C,q,\estimatorOrig)$ sample using identical instructions. We then aggregated verdicts with a conservative supermajority rule: an instance was labeled \emph{Likely Error} only if at least 4 of 5 runs flagged $\estimatorOrig$ as incorrect in a clinically significant way; otherwise it was labeled \emph{Passed}.

\paragraph{Clinician spot-check.}
To assess if the automated audit's flags were clinically sensible, a physician author (AG) spot-checked seven cases flagged as \emph{Likely Error} within his clinical domain. For each selected case, the physician reviewed the patient context, the calculator question, the original benchmark label, and the auditor's written critique, and recorded whether the critique reflected a clinically meaningful label problem.

\subsection{Phase 2: Independent recomputation of reference labels}
\label{sec:phase2_relabel}

\paragraph{Objective and Scope.}
Phase 2 proposes revised reference labels by \emph{independently} recomputing each score from first principles, without access to the original benchmark label or its derivation artifacts. This phase aims to (i) construct a set of high-confidence updated labels for evaluation and downstream use, and (ii) quantify disagreement between independently recomputed labels and the original labels to triage cases for clinician adjudication in Phase~3. We applied the Phase 2 relabeling pipeline to the full MedCalc-Bench test split ($n=1{,}047$) \emph{and} to a stratified random sample of the training split ($n=5{,}183$, ${\sim}50\%$ of the original 10{,}053-instance training set), stratified by calculator question type $q$ so that each of the 55 calculators is represented in approximately the same proportion as in the full training set.

\paragraph{Independent labeling agent with comprehensive tools access.}
We implemented a separate relabeling agent (Gemini 2.5 Pro) configured as a tool-augmented clinical calculator. For each instance, the agent received only the patient context and calculator question $(C,q)$ (not seeing $\estimatorOrig$ nor any released extraction/aggregation metadata), to reduce anchoring to the original pipeline. The agent had access to (i) a Google Search API to retrieve authoritative calculator definitions and scoring rules (e.g., MDCalc and guideline documents), and (ii) a Python code execution sandbox for arithmetic and unit conversions. 

The agent was instructed to (1) identify the appropriate scoring rule, (2) extract all relevant criteria from $C$, (3) map extracted evidence to score components, and (4) compute the final value. When required inputs were missing or the question could not be resolved clinically from the provided context, the agent was instructed to abstain (``$\nullnotation$'') rather than infer or hallucinate. For reliable parsing, the agent was required to enclose its final output in a strict regex-readable format (a scalar numeric value, a datetime, or ``$\nullnotation$''). The complete system and user prompts are provided in Supplementary Appendix \S\ref{app:prompts}.

\paragraph{Supermajority-Based Labeling.} To ensure stability and reliability, each instance is processed by five independent copies of the Gemini agent at a temperature of 1. We employ a supermajority voting rule: a ``high-confidence'' label ($\estimatorNew$) is recorded if at least four out of the five independent runs agree on the output type and value.\footnote{To balance stability with data retention, we also retain instances with a 3/5 supermajority if at least one of the remaining two outputs falls within a proximity threshold: $\pm5\%$ for numeric values or $\pm1$ day for dates. This ``near-consensus'' logic reduces unnecessary deferrals from minor variations in agent outputs.} Instances failing to reach this quorum are marked as ``Deferred'' and excluded from the high-confidence set.
For real-valued numeric outputs, agreement for the majority is defined after rounding to two decimal places; for datetimes and categorical outputs, an exact match is required.

\paragraph{Automated Triage for Physician Review.} To identify high-impact label failures for targeted clinician recomputation in Phase 3, we compare $\estimatorNew$ against $\estimatorOrig$ for each instance. We define a symmetric discrepancy metric, $\textsf{rel.err}$, for the two label estimators:
\begin{equation}\label{eq:rel-err}
\textsf{rel.err}(\estimatorOrig, \estimatorNew) := \frac{|\estimatorOrig - \estimatorNew|}{\max(|\estimatorOrig|, |\estimatorNew|)}
\end{equation}
Following the $\pm5\%$ tolerance used by MedCalc-Bench for grading model outputs, an instance is flagged as a ``Likely Error'' if $\textsf{rel.err}(\estimatorOrig, \estimatorNew) > 0.05$. Cases where either label is N/A or datetime are handled respectively by exact match and specific time-threshold differences.

\subsection{Phase 3: Targeted physician adjudication}
\label{sec:phase3-physician}

\paragraph{Objective.}
Phase 3 concentrates independent physician recomputation and review efforts on the subset of test instances with the largest disagreement between the original label ($\estimatorOrig$) and the Phase 2 recomputed label ($\estimatorNew$), thus providing a targeted adjudication.

\paragraph{Triage and selection.} Physician authors (DT, AG, NK) first screened calculator questions $q$ for respective domain fit. Author JY ranked test instances by disagreement between $\estimatorOrig$ and $\estimatorNew$, treating any abstention ($\nullnotation$) by either label as maximal disagreement, and then sampled 50 instances from the ranked list, assigning each instance to a physician aligned with the score type.
Each selected instance was reviewed once by a single physician.

\paragraph{Single-blind physician recomputation.}
For each assigned instance, the physician received only the patient context and question $(C,q)$ and independently recomputed the score from first principles in most of these, without seeing $\estimatorOrig$ or $\estimatorNew$ (single-blind to reduce anchoring).
Physicians could consult external references (e.g., MDCalc or guideline documents) and could abstain ($\nullnotation$) when the score was not clinically computable from the context.
Physicians optionally recorded brief free-text comments (e.g., missing inputs, ambiguity, guideline/version assumptions).
We denote a physician-computed label by $y^{\star}$.

\paragraph{Agreement criteria, error metrics, and uncertainty.}
We compare $\estimatorOrig$ and $\estimatorNew$ to $y^{\star}$ using (i) an agreement indicator and (ii) symmetric mean absolute percentage error (sMAPE, \cite{armstrong1985}) for answerable numeric cases.
For real-valued outputs, agreement is defined as relative deviation within $\pm 5\%$ (i.e., $|\hat y-y^{\star}|/|y^{\star}|\le 0.05$).
For integer-valued ordinal scores, agreement is defined as $|\hat y-y^{\star}|\le 1$ when $y^{\star}<20$.
For abstentions, agreement requires an exact match (both $y^{\star}$ and $\hat y$ equal $\nullnotation$).
For $N$ instances, sMAPE is computed as:
\begin{equation}\label{eq:smape}
\mathrm{sMAPE}(\{y^{\star}_i\}_{i=1}^{N},\{\hat y_i\}_{i=1}^{N})=\frac{100\%}{N}\sum_{i=1}^{N}\frac{2|y^{\star}_{i}-\hat y_i|}{|y^{\star}_{i}|+|\hat y_i|}.
\end{equation}
We estimate 95\% confidence intervals for sMAPE via bootstrapping 10{,}000 resamples.

\paragraph{Post-hoc label finalization and qualitative analysis.}
We used physician recomputations to finalize labels on the reviewed subset, yielding $\estimatorFinal$ where $\estimatorFinal=y^{\star}$ for physician-reviewed instances and $\estimatorFinal=\estimatorNew$ otherwise. For instances where $\estimatorFinal \neq y^{\star}$, we examine reasoning traces of the Gemini labeling agent, and if a systematic bias is identified relative to the physician's ground truth (e.g. scoring criteria of some $q_i$), we analogously revise $\estimatorNew$ labels for unreviewed $(C,q_i)$ instances. We also perform a post-hoc qualitative coding of physicians' free-text comments to summarize recurrent causes of disagreement. We release both $\estimatorNew$ (pre-adjudication) and $\estimatorFinal$ (post-adjudication) with triage metadata.

\subsection{Controlled Study of Label Quality's Downstream Impact}\label{sec:rl}
To determine if reference label inaccuracies are just benign ``white noise'' or a directional distortion that misaligns model behavior, we performed a controlled reinforcement learning (RL) experiment. This simulates the potential consequence when an AI researcher optimizes for biased reference labels that reflect synthetic artifacts rather than clinical reality.

We finetuned two identical copies of \href{https://huggingface.co/Qwen/Qwen3-8B}{{Qwen3-8B}}, an open-weights LLM with 4.6 million downloads from Hugging Face in August 2025 \cite{qwen3}, using \emph{Group Relative Policy Optimization (GRPO)}, a modern RL algorithm widely used in LLM alignment since its introduction in the DeepSeek-R1 work \cite{grpo, deepseek-r1}. Our setting admits a simplified, on-policy formulation of GRPO; for the reader’s convenience, we provide a self-contained derivation in Supplementary Appendix~\S\ref{sec:model-alignment-mdp}. The key property for our experiment is that the reference label ($y^\star$, surrogated in practice by $\hat y$) directly specifies a rule-based reward function; changing $\hat y$ is therefore functionally equivalent to changing the model’s implicit alignment objective.

To isolate the causal effect of reference label choice, all training factors, including the base model, prompt templates, hyperparameters, and compute budget, were held constant while varying only the source of the training reward:
\begin{itemize}
    \item \textbf{Control Arm (Original Labels):} One copy of Qwen3-8B was trained to maximize agreement with the original MedCalc-Bench reference labels ($\estimatorOrig$).
    \item \textbf{Treatment Arm (Updated Labels):} Another copy of Qwen3-8B was trained to maximize agreement with our Phase 2 independently recomputed labels ($\estimatorNew$).
\end{itemize}
Full experimental details, including hyperparameters, hardware, and the sampling and evaluation protocol, are given in Supplementary Appendix~\S\ref{app:RL}.

Both models were trained on 4{,}593 training instances. We evaluate the resulting checkpoints along three dimensions: (1) accuracy on the 50 physician-labeled test instances ($y^\star$) from Phase 3, which serves as our primary measure; (2) accuracy on the full 887-instance held-out test set, graded against $\estimatorNew$; and (3) accuracy on three additional medical evaluation sets (MedQA and two recent MedCalc-Bench revisions) to test whether training-label effects generalize to related tasks. The primary result is reported in \S\ref{results:RL}; full details are in Supplementary Appendix~\S\ref{app:RL}.

\subsection*{Data and Code Availability}
We have posted the raw outputs of Phases 1 and 2 workflows, high-confidence recomputed labels of MedCalc test set instances and train set instances obtained in Phase 2, physician adjudication results in Phase 3, and code to reproduce the RL experiment. To access, visit the GitHub repository: \href{https://github.com/junzeye/validate-medcalc-labels}{https://github.com/junzeye/validate-medcalc-labels}.
\section{Results}

\subsection{Phase 1: Automated audit of original labels}\label{sec:phase1-results}

Running five independent copies of the auditor agent on MedCalc's test set (n=1{,}047) flagged 279 (26.6\%) instances as as \emph{Likely Error}, i.e. marked as \emph{Suspicious} by at least 4 out of 5 auditor reviews. Flagged issues were distributed broadly across the different questions $q$, covering 40 out of the total 55 distinct scoring questions in the test set.

\begin{figure}[htbp]
    \centering
    \includegraphics[width=1\linewidth]{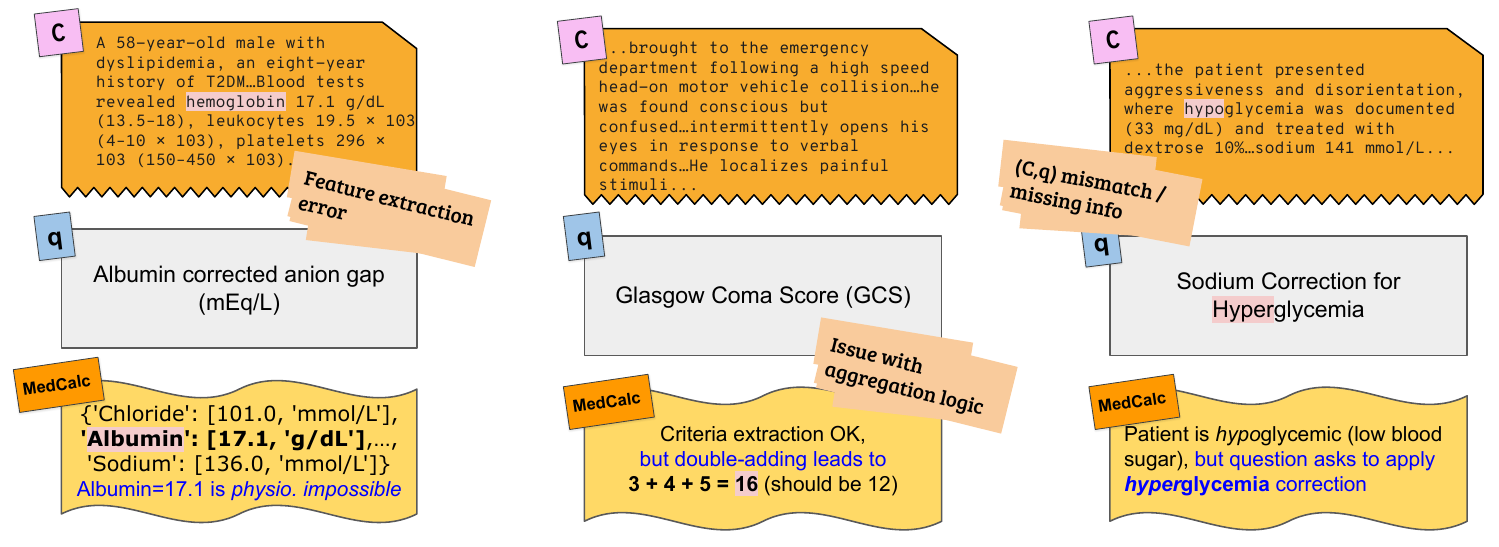}
    \caption{\emph{Representative error types.} (a) Feature extraction error: GPT-4 might have confused ``hemoglobin'' with ``albumin'', extracting a value that is physiologically impossible; (\textbf{b}) Incorrect aggregation logic: an incorrect Python code for Glasgow Coma Scale aggregation that double-counts a feature value, inflating $\estimatorOrig$; (\textbf{c}) $q$ is not answerable given $C$: a Sodium correction for \emph{hyper}glycemia inappropriately applied to a \emph{hypo}glycemic patient.}
    \label{fig:pipeline1-out2}
\end{figure}

\paragraph{Qualitative error taxonomy.}
Manual inspection of flagged instances reveals three recurring failure modes in the original ground truth derivation: (a) \emph{feature extraction errors}, (b) \emph{incorrect aggregation logic}, i.e. flawed scoring rules, and (c) \emph{$q$ is not answerable given $C$}. Figure~\ref{fig:pipeline1-out2} illustrates one example of each. In the case of (c), it would be more reasonable to reward the LLM for \emph{abstaining} from answering the impossible and responding with $\nullnotation$ (``Not Available''), instead of hallucinating an answer. Moreover, instead of throwing away all these instances, we suggest that the benchmark recycle them by keeping $(C,q)$ but revising the ground truth label from $\estimatorOrig$ to $\nullnotation$ (see Supplementary Appendix \S\ref{app:na-examples} for two examples). Correspondingly, the benchmark's inference prompt could acknowledge that there are instances where the correct answer is to abstain. A safely useful LLM system, we believe, should act as a triage agent that seeks human review for out-of-distribution $(C,q)$ instances, while strictly limiting abstention on normal cases to preserve the efficiency gains of automation. 

\paragraph{Clinician spot-check feedback.}
A physician author (AG) independently selected and reviewed 7 \emph{Suspicious} instances related to their clinical specialty. For each instance, the physician read both $(C,q,\estimatorOrig)$ and the auditor agents' critique of $\estimatorOrig$. They agreed with the error identification in 7/7 reviews, suggesting the clinical utility of running a larger-scale MedCalc label recomputation in Phase 2.

\subsection{Phase 2: Independent recomputation of labels}\label{sec:phase2-results}

\paragraph{Coverage and abstentions.}
The supermajority-based labeling approach discussed above (see Methods \S\ref{sec:phase2_relabel} for full details) yielded high-confidence labels for 887 out of 1{,}047 MedCalc test instances (85\%) and for 4{,}593 out of 5{,}183 sampled train instances (89\%).
In these high-confidence labels, the abstention rate, i.e. how often the finalized label is $\nullnotation$, was 7.4\% on the test set and 5.7\% on the train set.

\paragraph{Estimated rate of likely mislabels.}
Among the 887 test instances with high-confidence $\estimatorNew$, 220 (24.8\%) differ from the original label by more than the $\pm 5\%$ (formally defined in Methods Eq.~\eqref{eq:rel-err}), and another 66 (7.4\%) are flagged as $\nullnotation$. Together, these imply that at least 27.3\% of the 1{,}047 original test instances are likely mislabeled. This is plausibly an underestimate: the 160 test instances for which the labeling agents could \emph{not} reach a supermajority are deferred  because they are hard cases where both $\estimatorOrig$ and $\estimatorNew$ are unreliable, and the true mislabel rate could be substantially higher once these are accounted for.

\paragraph{Cross-phase consistency check.}
The close correspondence between Phase 1's audit-based error rate (26.6\%) and Phase 2's recomputation-based estimate (27.3\%) provides a sanity check. These phases differ meaningfully in their setup: Phase 1 audits labels with full metadata visibility, while Phase 2 recomputes without the anchoring. Still, they arrive at similar error estimates. While not constituting direct validation, this consistency suggests that the detected quality issues are robust to variations in prompting and workflow design.

\subsection{Phase 3: Physician adjudication of disagreements}\label{sec:phase3-results}

\begin{figure}[htbp]
    \centering
    \includegraphics[width=0.95\linewidth]{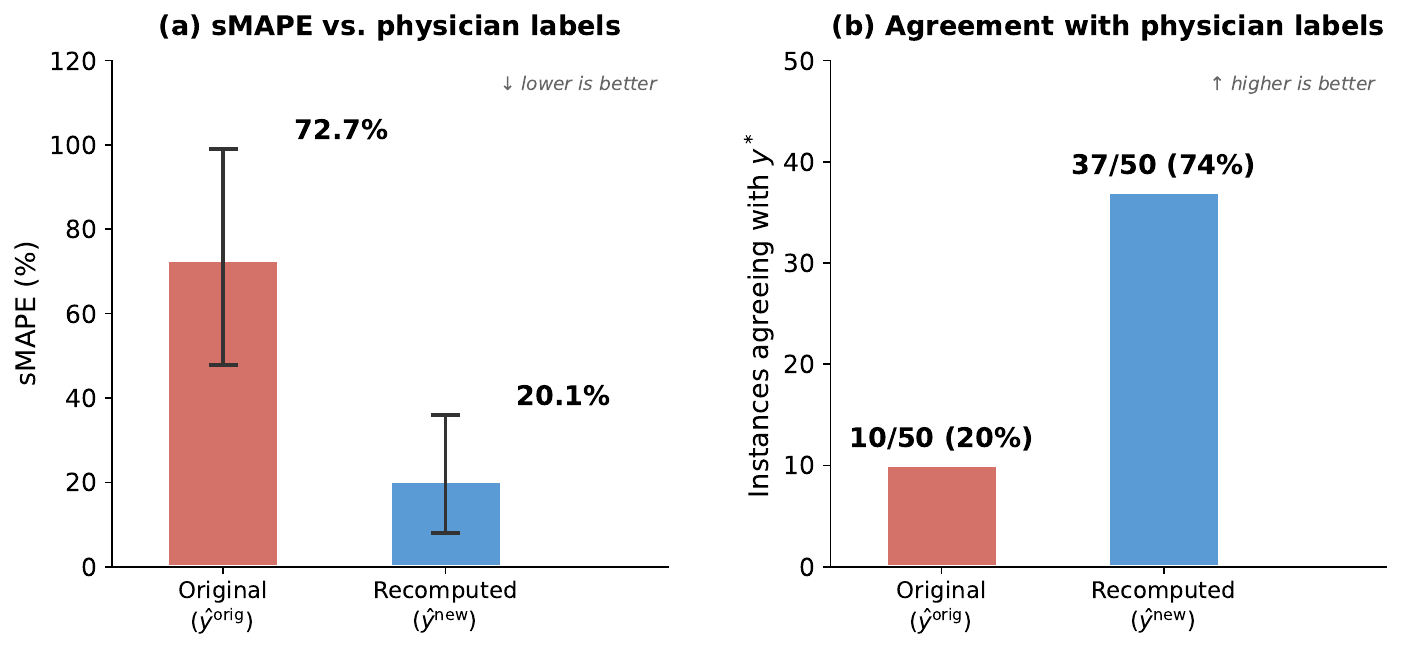}
    \caption{Original MedCalc-Bench labels ($\estimatorOrig$) and our recomputed labels ($\estimatorNew$), compared against 50 physician-computed labels ($y^\star$). \textbf{(a)} sMAPE (symmetric Mean Absolute Percentage Error) measures the average relative deviation from physicians' labels; error bars show bootstrap 95\% confidence intervals. \textbf{(b)} Agreement counts how many labels match the physician label up to a reasonable threshold ($\pm5\%$ for numeric, exact match for abstentions).}
    \label{fig:phase3-result}
\end{figure}

Figure~\ref{fig:phase3-result} compares the accuracy of $\estimatorOrig$ and $\estimatorNew$ on 50 test instances with physician-computed labels ($y^\star$). $\estimatorNew$ agrees with $y^\star$ 37 out of 50 times (74\%; 95\% CI, 60--84\%), compared to $\estimatorOrig$'s 10 out of 50 (20\%; 95\% CI, 11--33\%). Beyond binary agreement, the recomputed labels drive sMAPE down from 72.7\% (95\% CI, 47.9--99.1) to 20.1\% (95\% CI, 8.0--35.9), meaning that $\estimatorNew$ typically lies much closer to physicians' numerical assessments. Although the sample size is modest, these results provide a targeted reality check that the relabeled $\estimatorNew$'s are more aligned with expert judgment on the contentious instances triaged by Phase~2.

A conservative extrapolation to the full test set and a post-hoc qualitative analysis of physician comments are provided in Supplementary Appendix \S\ref{app:sec33-math}.

\begin{figure}[t]
  \centering
  \begin{subfigure}[b]{0.48\textwidth}
    \centering
    \includegraphics[width=\textwidth]{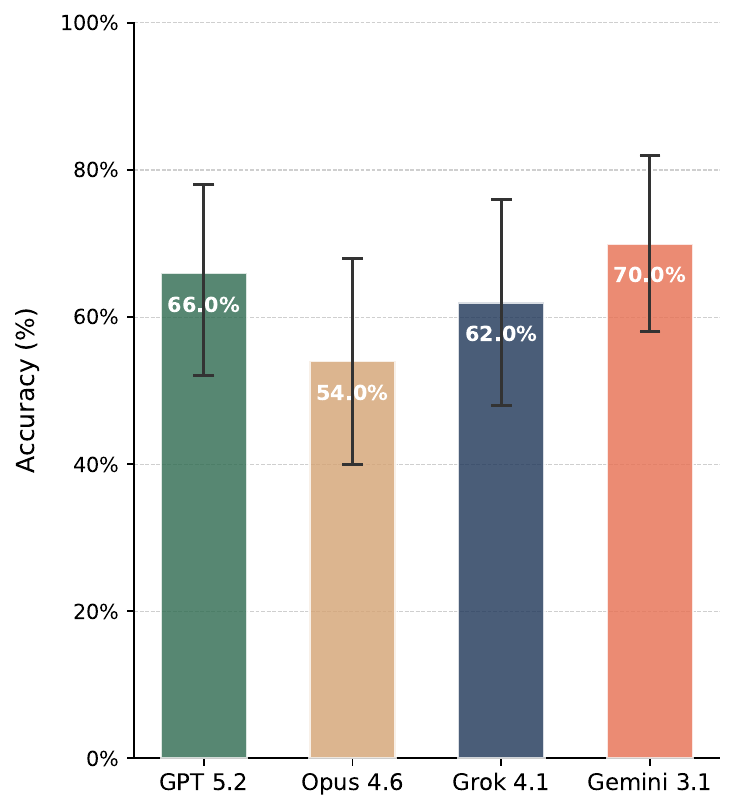}
    \caption{Physician-produced labels ($N=50$)}
    \label{fig:eval-50ex}
  \end{subfigure}
  \hfill
  \begin{subfigure}[b]{0.48\textwidth}
    \centering
    \includegraphics[width=\textwidth]{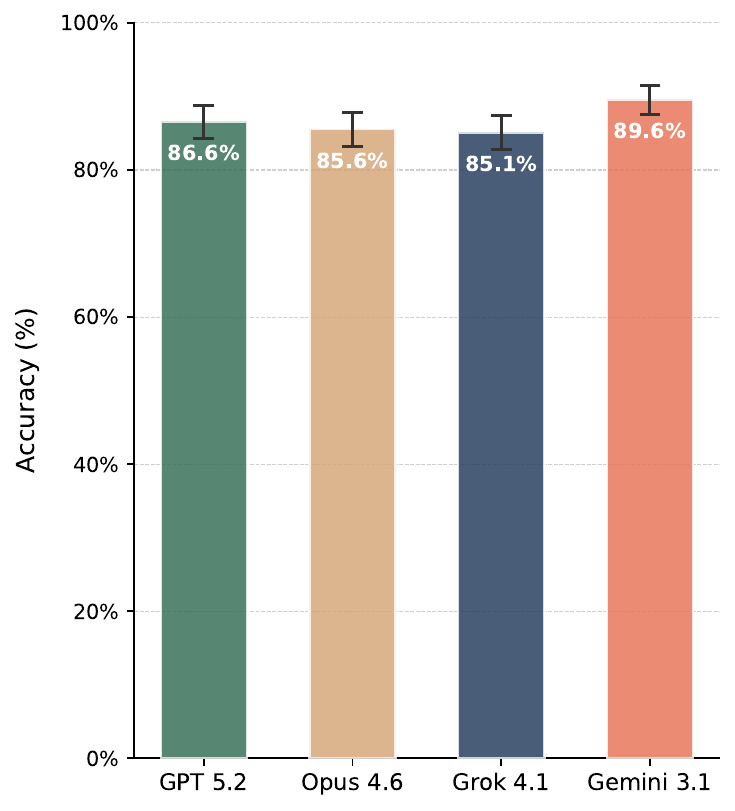}
    \caption{Full-pipeline labels ($N=887$)}
    \label{fig:eval-full}
  \end{subfigure}
  \caption{Frontier LLM accuracy ($\pm 95\%$ CI) on MedCalc-Bench test split under our newly curated reference labels. Models are evaluated with a uniform prompt template and built-in tool use (web search + code execution) under high reasoning effort. \textbf{(a)} Accuracy against \emph{physician-computed labels} for a subset of 50 test instances flagged for manual review in Phase 3. \textbf{(b)} Accuracy on all 887 test instances our stewardship pipeline relabeled, combining labels curated from Phase 2 \& 3.}
  \label{fig:frontier-llm-on-our-curated}
\end{figure}

\paragraph{Evaluating today's LLMs with our curated labels.}\label{results:eval-frontier} Figure \ref{fig:frontier-llm-on-our-curated} compares several frontier models in February 2026, evaluated against reference labels we curated in our three-phase pipeline. Figure~\ref{fig:eval-50ex} shows model accuracy on the 50-instance subset for which we have physician-computed labels ($y^\star$). Performance on this subset is expectedly lower than on the full test set, since these are the instances flagged in Phase 2 as the hardest and most contentious cases. Figure~\ref{fig:eval-full} reports accuracy on the full 887 instances for which our pipeline produced high-confidence recomputed labels. Because physician ground truth is unavailable for most of these instances, the comparison here uses $\estimatorNew$ as a proxy for $y^\star$, which is justified by the Phase 3 evidence that $\estimatorNew$ aligns substantially more closely with physician judgment than $\estimatorOrig$ does (Figure~\ref{fig:phase3-result}). Accuracies in Figure~\ref{fig:eval-full} are therefore measured against a pseudo-ground-truth, but the larger sample size yields tighter confidence intervals.

We note that for the same 887 instances, the original MedCalc-Bench labels are also available as a reference. Grading the same model outputs against the original labels yields substantially lower accuracies for all four models ($65.7\%$, $69.4\%$, $62.5\%$, $67.8\%$, in the same order as Figure~\ref{fig:eval-full}). This drop is not a difference in the models, but a direct consequence of label errors in the original benchmark, providing further evidence that the original labels meaningfully \emph{underestimate} the capability of today's frontier LLMs on medical score computation. These lower accuracy figures are consistent with numbers reported by industry: in Anthropic's ``Claude for Healthcare'' launch \cite{anthropic_press}, our independent reproduction analysis (Supplementary Appendix~\S\ref{app:claude}) suggests that these figures were likely obtained using the original v1.0 labels, further illustrating how label noise in the benchmark can understate model performance even in high-profile industry evaluations.

\subsection{Effect of Label Choice on Downstream RL Training}\label{results:RL}

\begin{figure}[htbp]
    \centering
    \includegraphics[width=1.0\linewidth]{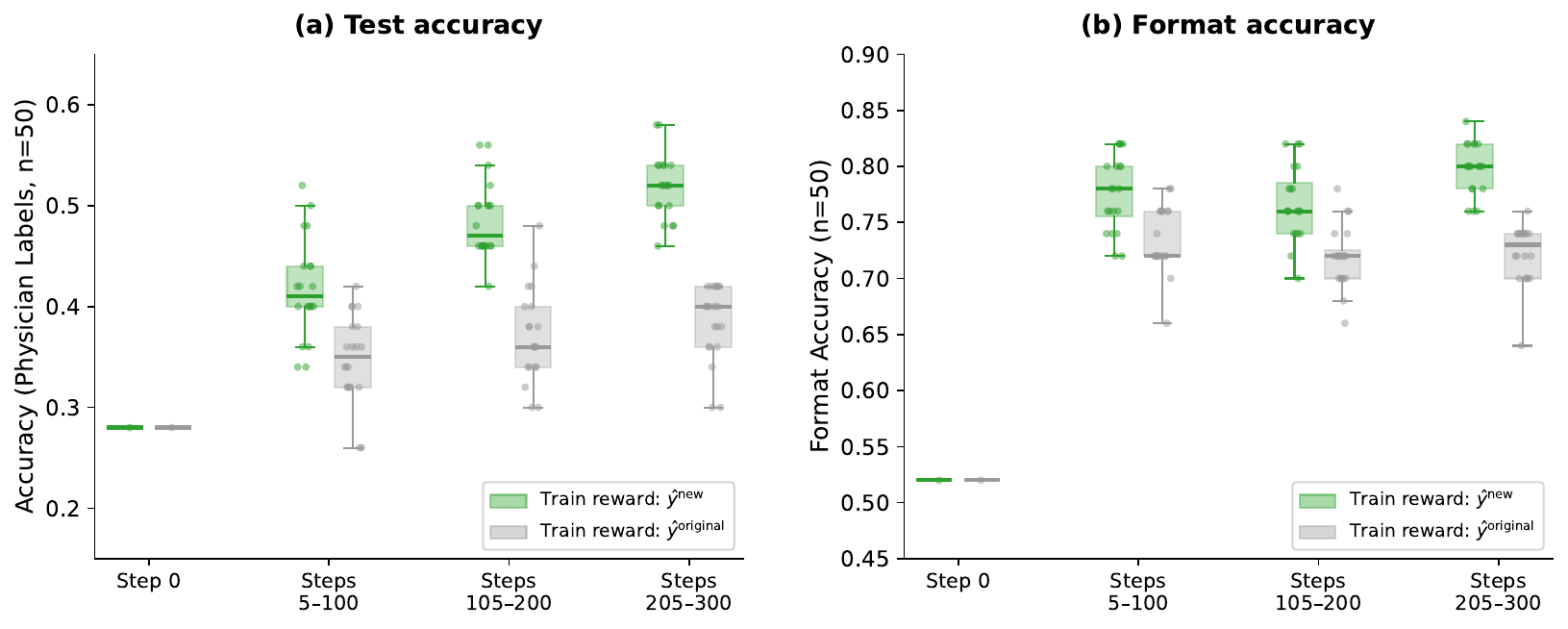}
    \caption{RL-trained model performance on 50 physician-labeled test instances ($y^\star$). Two identical copies of Qwen3-8B are trained via GRPO, differing only in whether training rewards are derived from the original labels ($\estimatorOrig$, grey) or our recomputed labels ($\estimatorNew$, green). Each box summarizes the distribution of per-checkpoint accuracies within the indicated step range; individual checkpoints (evaluated every 5 steps) are overlaid as dots. \textbf{(a)} Test accuracy against physician labels. \textbf{(b)} Format accuracy (whether the model's output is parseable as a valid score).}
    \label{fig:rl-physician-eval}
\end{figure}

To test whether label quality affects not only evaluation but also model training, we evaluate the two RL-trained Qwen3-8B checkpoints (one trained on $\estimatorOrig$, the other on $\estimatorNew$; see Methods \S\ref{sec:rl}) on the 50 physician-labeled test instances from Phase 3. Both models start from the same base checkpoint and achieve identical accuracy before training (28\%). Figure~\ref{fig:rl-physician-eval} shows how their performance on physician-labeled instances evolves over the course of training. As training progresses, a clear gap emerges: during the final phase of training (steps 205--300), the model trained on $\estimatorNew$ achieves a mean accuracy of 51.9\% (95\% CI, 50.5--53.3) versus 38.4\% (95\% CI, 36.7--39.9) for the model trained on $\estimatorOrig$, a difference of approximately 13.5 percentage points attributable solely to the choice of training reward labels.

When evaluated on the larger 887-instance held-out test set (graded against $\estimatorNew$), the gap narrows to $+8.7$ percentage points ($71.4\%$ vs $62.6\%$; see Appendix~\S\ref{app:RL} and Figure~\ref{fig:rl-result-main} for the full training dynamics). The smaller gap on the full test set is expected: the 50 physician-labeled instances were selected precisely because they had the highest disagreement between $\estimatorOrig$ and $\estimatorNew$, so the effect of label choice is concentrated there. On the remaining instances, where the two label sets largely agree, the choice of training reward matters less, and the test-wide average reflects a mixture of both regimes. We also evaluate both checkpoints on three additional medical evaluation sets (MedQA and two recent MedCalc-Bench revisions). The advantage of training on $\estimatorNew$ persists, though the effect is smaller (approximately 1 percentage point; $p=0.001$). Importantly, training on higher-quality labels does not degrade performance on any of these related tasks; the improvement on the primary task comes without a trade-off elsewhere (Appendix~\S\ref{app:RL-ood}).

All recomputed labels, physician adjudication results, raw pipeline outputs, and code to reproduce the RL experiments are publicly available at \href{https://github.com/junzeye/validate-medcalc-labels}{https://github.com/junzeye/validate-medcalc-labels}.
\section{Discussion}

In this work, we show that label quality in LLM-assisted benchmarks can meaningfully change both evaluation conclusions and model alignment outcomes. LLM assistance in benchmark construction is not a shortcut that can be avoided; it is increasingly a necessity. As models become more capable, benchmarks must grow harder to remain informative, and the expert labor required to produce ground-truth labels at scale, particularly in domains like medicine where that labor comes from clinicians with limited availability, makes fully human annotation impractical. LLM-assisted labeling is therefore likely to become more common, not less. The practical implication is not that a benchmark must be ``perfect’’ at release, but that it should be maintained as a living document, with labels, assumptions, and scoring criteria revisited over time: even though models used for labeling continue to improve, they still hallucinate and make errors, and as the broader community engages with a benchmark, new issues surface that warrant correction. More capable auditing tools and broader community feedback together make ongoing stewardship both necessary and increasingly feasible.

A binding constraint in operationalizing this kind of stewardship is the finitude of domain experts’ attention. GPU hours and API calls are elastic to scale; expert hours are not. Any practical stewardship protocol must fit into an already stretched professional environment and should minimize disruption to routine workflows. We therefore frame stewardship as a hybrid oversight system: automated audit and independent recomputation provide broad screening at low marginal cost, and expert adjudication is reserved for a small set of high-disagreement cases. The goal is not to replace expert judgment, but to concentrate it where it is most likely to change labels and improve alignment. While we demonstrate this approach in medicine, the same design principle applies to any domain where expert labels are expensive and benchmark accuracy is safety-relevant, including legal reasoning, materials science, and financial risk assessment.

Our qualitative analysis (Supplementary Appendix~\S\ref{app:sec33-math}) reveals that a substantial share of label disagreement stems from context underspecification: benchmark instances that omit required inputs or leave key qualifiers implicit. In these settings, forcing a numeric answer is unsafe. Explicit abstention options, and evaluation rules that credit abstention when the answer is not computable, should be treated as a safety guardrail rather than an edge-case exception.

Beyond MedCalc-Bench, we view benchmark stewardship as one concrete way for academia and the broader AI community to help shape the trajectory of the next generation of LLMs. Frontier model development is increasingly concentrated in a small number of well-resourced AI labs; this concentration can reduce direct feedback from diverse real-world settings, affecting relevance and accuracy. Public benchmarks therefore play a major role in setting norms: they define the state of the art and train the intuitions of engineers who optimize against shared test sets. When public benchmarks contain uncorrected errors, they may inadvertently obscure the true capabilities of frontier models, turning accountability tools into sources of measurement ambiguity.

Label quality matters not only for measurement but also for shaping the next generation of language models. Benchmarks are increasingly reused for model selection and post-training; in this role, the yardstick becomes a teacher. Our controlled RL experiment isolates this effect by holding the training pipeline fixed and changing only the reward labels, producing a 13.5 percentage-point gap in accuracy on physician-labeled instances. Importantly, training on higher-quality labels does not come at the cost of performance on related tasks: the advantage persists, albeit at a smaller magnitude, across additional medical evaluation sets. This finding connects to a broader concern in the AI community: when LLM-generated outputs are fed back into the training pipeline, whether as benchmark labels, synthetic training data, or reward signals, errors can compound across generations. This feedback loop has been studied under the heading of ``model collapse,’’ where recursively generated data can degrade future models if incorporated without adequate quality controls \cite{gerstgrasser2024modelcollapse,shumailov2024collapse}. Our results provide concrete evidence of a related mechanism in the benchmark setting: LLM-generated labels that are treated as static ground truth can silently bias the models trained against them.

Several limitations should be noted when interpreting these results. First, our physician ground truth comprises 50 instances, which is sufficient to detect the large effects we report but lacks statistical power to detect smaller differences. Second, we treat physician labels as ground truth, but clinicians themselves can make calculation errors or disagree on ambiguous cases. Ideally, each instance would be independently labeled by multiple clinicians from different institutions, and their responses aggregated to reduce noise in the reference standard; our single-physician design does not capture this inter-rater variability (Supplementary Appendix~\S\ref{sec:estimation-lens} discusses the implications of treating ground truth as a point estimate versus a distribution). Third, while our five-agent consensus mechanism in Phases 1 and 2 reduces stochastic noise, the agents are not fully independent: they share the same underlying model and prompting strategy, so systematic biases (e.g., a consistent misinterpretation of a scoring rule) would not be caught by majority voting. Finally, the RL experiment uses a single 8B-parameter model; larger models may exhibit different generalization behavior, and testing this is an important direction for future work.

Taken together, these findings argue for treating LLM-assisted benchmarks as living documents: transparent versioning, documented assumptions, explicit representation of ``unknown,’’ and hybrid oversight that focuses expert effort where it matters most. This is a pragmatic path to preserving the public-good role of benchmarks, so that they remain trustworthy yardsticks for evaluation, and safer teachers when they are reused to shape the models we deploy.

\section*{Acknowledgments}
Some of the computing for this project was performed on the Stanford Marlowe GPU cluster \cite{marlowe}. We thank Stanford University and Stanford Data Science for providing the computational resources at a subsidized rate that made these research results possible.

\bibliographystyle{plainnat}
\bibliography{references}

@inproceedings{northcutt2021labelerrors,
  title={Pervasive Label Errors in Test Sets Destabilize Machine Learning Benchmarks},
  author={Northcutt, Curtis G. and Athalye, Anish and Mueller, Jonas},
  booktitle={Advances in Neural Information Processing Systems},
  volume={34},
  year={2021},
  publisher={Curran Associates, Inc.}
}

@article{nahum2024labelerrors,
  title={Are {LLMs} Better than Reported? {D}etecting Label Errors and Mitigating Their Effect on Model Performance},
  author={Nahum, Omer and Calderon, Nitay and Keller, Orgad and Szpektor, Idan and Reichart, Roi},
  journal={arXiv preprint arXiv:2410.18889},
  year={2024}
}

@article{medqa,
  title={What disease does this patient have? a large-scale open domain question answering dataset from medical exams},
  author={Jin, Di and Pan, Eileen and Oufattole, Nassim and Weng, Wei-Hung and Fang, Hanyi and Szolovits, Peter},
  journal={Applied Sciences},
  volume={11},
  number={14},
  pages={6421},
  year={2021},
  publisher={MDPI}
}

@article{medagentbench,
  title={MedAgentBench: a virtual EHR environment to benchmark medical LLM agents},
  author={Jiang, Yixing and Black, Kameron C and Geng, Gloria and Park, Danny and Zou, James and Ng, Andrew Y and Chen, Jonathan H},
  journal={NEJM AI},
  volume={2},
  number={9},
  pages={AIdbp2500144},
  year={2025},
  doi={10.1056/AIdbp2500144},
  publisher={Massachusetts Medical Society}
}

@misc{anthropic_youtube,
  author       = {{Anthropic}},
  title        = {The Briefing: Healthcare and Life Sciences},
  howpublished = {YouTube video (livestream / on-demand recording)},
  year         = {2026},
  month        = jan,
  day          = {12},
  url          = {https://www.youtube.com/watch?v=UXyVMGAFLAs},
  urldate      = {2026-01-17},
  note         = {At 23:09-24:16, the presenter discusses Claude model series' performance on MedCalc-Bench and MedAgentBench.}
}

@misc{anthropic_press,
  author       = {{Anthropic}},
  title        = {Advancing Claude in healthcare and the life sciences},
  year         = {2026},
  month        = {1},
  day          = {11},
  howpublished = {Anthropic (News / Announcements)},
  url          = {https://web.archive.org/web/20260112115216/https://www.anthropic.com/news/healthcare-life-sciences},
  note         = {Archived by the Internet Archive Wayback Machine. See Figure 1 of their press release.}
}

@article{lace,
  title={Derivation and validation of an index to predict early death or unplanned readmission after discharge from hospital to the community},
  author={Van Walraven, Carl and Dhalla, Irfan A and Bell, Chaim and Etchells, Edward and Stiell, Ian G and Zarnke, Kelly and Austin, Peter C and Forster, Alan J},
  journal={Can Med Assoc J},
  volume={182},
  number={6},
  pages={551--557},
  year={2010},
  publisher={CMAJ},
  doi={10.1503/cmaj.091117}
}

@article{chung2025,
  title={Verifying Facts in Patient Care Documents Generated by Large Language Models Using Electronic Health Records},
  author={Chung, Philip and Swaminathan, Akshay and Goodell, Alex J and Kim, Yeasul and Momsen Reincke, S and Han, Lichy and Deverett, Ben and Sadeghi, Mohammad Amin and Ariss, Abdel-Badih and Ghanem, Marc and others},
  journal={NEJM AI},
  volume={3},
  number={1},
  pages={AIdbp2500418},
  year={2025},
  doi={10.1056/AIdbp2500418},
  publisher={Massachusetts Medical Society}
}

@misc{google_ai_param,
  author       = {{Google Cloud}},
  title        = {Experiment with parameter values | Generative AI on Vertex AI},
  year         = {2025},
  url          = {https://docs.cloud.google.com/vertex-ai/generative-ai/docs/learn/prompts/adjust-parameter-values},
  note         = {Accessed: 2025-12-30}
}

@inproceedings{
gerstgrasser2024modelcollapse,
title={Is Model Collapse Inevitable? Breaking the Curse of Recursion by Accumulating Real and Synthetic Data},
author={Matthias Gerstgrasser and Rylan Schaeffer and Apratim Dey and Rafael Rafailov and Tomasz Korbak and Henry Sleight and Rajashree Agrawal and John Hughes and Dhruv Bhandarkar Pai and Andrey Gromov and Dan Roberts and Diyi Yang and David L. Donoho and Sanmi Koyejo},
booktitle={First Conference on Language Modeling},
year={2024},
url={https://openreview.net/forum?id=5B2K4LRgmz},
publisher={PMLR}
}

@article{shumailov2024collapse,
  title        = {AI models collapse when trained on recursively generated data},
  author       = {Shumailov, Ilia and Shumaylov, Zakhar and Zhao, Yiren and Papernot, Nicolas and Anderson, Ross and Gal, Yarin},
  journal      = {Nature},
  year         = {2024},
  volume       = {631},
  number       = {8022},
  pages        = {755--759},
  doi          = {10.1038/s41586-024-07566-y}
}

@article{afshar2025pragmatic,
  title={A pragmatic randomized controlled trial of ambient artificial intelligence to improve health practitioner well-being},
  author={Afshar, Majid and Ryan Baumann, Mary and Resnik, Felice and Hintzke, Josie and Gravel Sullivan, Anne and Wills, Graham and Lemmon, Kayla and Dambach, Jason and Mrotek, Leigh Ann and Quinn, Mariah and others},
  journal={NEJM AI},
  volume={2},
  number={12},
  pages={AIoa2500945},
  year={2025},
  doi={10.1056/AIoa2500945},
  publisher={Massachusetts Medical Society}
}

@article{medhelm,
  title={Holistic evaluation of large language models for medical tasks with MedHELM},
  author={Bedi, Suhana and Cui, Hejie and Fuentes, Miguel and Unell, Alyssa and Wornow, Michael and Banda, Juan M and Kotecha, Nikesh and Keyes, Timothy and Mai, Yifan and Oez, Mert and others},
  journal={Nature Medicine},
  pages={1--9},
  year={2026},
  doi = {10.1038/s41591-025-04151-2},
  publisher={Nature Publishing Group US New York}
}

@article{roeschl2025,
  title={Development of an LLM Pipeline Surpassing Physicians in Cardiovascular Risk Score Calculation},
  author={Roeschl, Tobias and Hoffmann, Marie and Unbehaun, Axel and Dreger, Henryk and Hindricks, Gerhard and Falk, Volkmar and Balicer, Ran and Tanacli, Radu and Hohendanner, Felix and Meyer, Alexander},
  journal={medRxiv},
  pages={2025--11},
  year={2025},
  publisher={Cold Spring Harbor Laboratory Press},
  doi={10.1101/2025.11.11.25340002}
}

@book{armstrong1985,
  title={Long-Range Forecasting: From Crystal Ball to Computer},
  author={Armstrong, J. Scott},
  year={1985},
  publisher={John Wiley \& Sons},
  address={New York},
  edition={2nd},
  isbn={978-0471030027},
  url={https://ssrn.com/abstract=666990}
}

@misc{utah_mme,
  author       = {{Utah Department of Health and Human Services}},
  title        = {{Opioid Oral Morphine Milligram Equivalent (MME) Conversion Factors}},
  year         = {2025},
  url          = {https://web.archive.org/web/20250801184557/https://medicaid-documents.dhhs.utah.gov/Documents/files/Opioid-Morphine-EQ-Conversion-Factors.pdf},
  note         = {Utah Medicaid. Archived from the original on August 1, 2025.}
}

@misc{maryland_mme,
  author       = {{Maryland Department of Health}},
  title        = {{PDMP Morphine Milligram Equivalents Fact Sheet}},
  year         = {2025},
  url          = {https://web.archive.org/web/20250831211406/https://health.maryland.gov/pdmp/Documents/Clinical%20Docs/MME%20Fact%20Sheet.pdf},
  note         = {Prescription Drug Monitoring Program. Archived from the original on August 31, 2025.}
}

@misc{cdc_opioids_2016,
  author       = {{CDC}},
  title        = {{Calculating Total Daily Dose of Opioids for Safer Dosage}},
  year         = {2016},
  url          = {https://web.archive.org/web/20221125005318/https://www.cdc.gov/drugoverdose/pdf/calculating_total_daily_dose-a.pdf},
  howpublished = {U.S. Department of Health and Human Services},
  note         = {Archived from the original on November 25, 2022.}
}

@misc{dowell2022cdc,
  author = {Dowell, Deborah and Ragan, Kathleen R. and Jones, Christopher M. and Baldwin, Grant T. and Chou, Roger},
  title = {{CDC} Clinical Practice Guideline for Prescribing Opioids for Pain --- {United States}, 2022},
  journal = {MMWR Recommendations and Reports},
  volume = {71},
  number = {3},
  pages = {1--95},
  year = {2022},
  month = {11},
  doi = {10.15585/mmwr.rr7103a1},
  url = {https://www.cdc.gov/mmwr/volumes/71/rr/rr7103a1.htm},
  publisher = {Centers for Disease Control and Prevention}
}

@misc{marlowe,
  author       = {Craig Kapfer and Kurt Stine and Balasubramanian Narasimhan
                  and Christopher Mentzel and Emmanuel Candes},
  title        = {Marlowe: Stanford's GPU-based Computational Instrument},
  year         = {2025},
  version      = {0.1},
  doi          = {10.5281/zenodo.14751899},
  publisher    = {Zenodo},
  note         = {Stanford Data Science report. \url{https://doi.org/10.5281/zenodo.14751899}}
}

@article{tierney2024ambient,
  author  = {Tierney, Aaron A. and Gayre, Gregg and Hoberman, Brian and Mattern, Brian and Ballesca, Mark and Kipnis, Patrick and Liu, Vincent and Lee, Keisa},
  title   = {Ambient Artificial Intelligence Scribes to Alleviate the Burden of Clinical Documentation},
  journal = {NEJM Catal Innov Care Deliv},
  year    = {2024},
  volume  = {5},
  number  = {3},
  pages   = {0404},
  doi     = {10.1056/CAT.23.0404}
}

@article{korom2025penda,
  author  = {Korom, Robert and Kiptinness, Sarah and Adan, Najib and Said, Kassim and Ithuli, Catherine and Rotich, Oliver and Kimani, Boniface and King'ori, Irene and Kamau, Stellah and Atemba, Elizabeth and Aden, Muna and Bowman, Preston and Sharman, Michael and Soskin Hicks, Rebecca and Distler, Rebecca and Heidecke, Johannes and Arora, Rahul K. and Singhal, Karan},
  title   = {{AI}-based Clinical Decision Support for Primary Care: A Real-World Study},
  journal = {arXiv preprint},
  year    = {2025},
  doi={10.48550/arXiv.2507.16947}
}

@article{lim2003curb65,
  title={Defining community acquired pneumonia severity on presentation to hospital: an international derivation and validation study},
  author={Lim, W.S. and van der Eerden, M.M. and Laing, R. and Boersma, W.G. and Karalus, N. and Town, G.I. and Lewis, S.A. and Macfarlane, J.T.},
  journal={Thorax},
  volume={58},
  number={5},
  pages={377--382},
  year={2003},
  publisher={BMJ Publishing Group},
  doi={10.1136/thorax.58.5.377}
}

@article{lip2010cha2ds2vasc,
  title={Refining clinical risk stratification for predicting stroke and thromboembolism in atrial fibrillation using a novel risk factor-based approach: the Euro Heart Survey on atrial fibrillation},
  author={Lip, Gregory Y.H. and Nieuwlaat, Robby and Pisters, Ron and Lane, Deirdre A. and Crijns, Harry J.G.M.},
  journal={Chest},
  volume={137},
  number={2},
  pages={263--272},
  year={2010},
  publisher={Elsevier},
  doi={10.1378/chest.09-1584}
}

@inproceedings{verl,
  title={Hybridflow: A Flexible and Efficient RLHF Framework},
  author={Sheng, Guangming and Zhang, Chi and Ye, Zilingfeng and Wu, Xibin and Zhang, Wang and Zhang, Ru and Peng, Yanghua and Lin, Haibin and Wu, Chuan},
  booktitle={Proceedings of the Twentieth European Conference on Computer Systems},
  pages={1279--1297},
  year={2025},
  doi={10.1145/3689031.3696075},
  publisher={ACM}
}

@article{drgrpo,
  title={Understanding r1-zero-like training: A critical perspective},
  author={Liu, Zichen and Chen, Changyu and Li, Wenjun and Qi, Penghui and Pang, Tianyu and Du, Chao and Lee, Wee Sun and Lin, Min},
  journal={arXiv preprint},
  year={2025},
  doi={10.48550/arXiv.2503.20783}
}

@inproceedings{ahmadian2024,
  title={Back to Basics: Revisiting REINFORCE-Style Optimization for Learning from Human Feedback in LLMs},
  author={Ahmadian, Arash and Cremer, Chris and Gall{\'e}, Matthias and Fadaee, Marzieh and Kreutzer, Julia and Pietquin, Olivier and {\"U}st{\"u}n, Ahmet and Hooker, Sara},
  booktitle={Proceedings of the 62nd Annual Meeting of the Association for Computational Linguistics (Volume 1: Long Papers)},
  pages={12248--12267},
  year={2024},
  publisher={Association for Computational Linguistics}
}

@misc{qwen3,
      title={Qwen3 Technical Report}, 
      author={Qwen},
      year={2025},
      eprint={2505.09388},
      archivePrefix={arXiv},
      primaryClass={cs.CL},
      url={https://web.archive.org/web/20250829140319/https://huggingface.co/Qwen/Qwen3-8B},
}

@inproceedings{
kumar25,
title={Training Language Models to Self-Correct via Reinforcement Learning},
author={Aviral Kumar and Vincent Zhuang and Rishabh Agarwal and Yi Su and John D Co-Reyes and Avi Singh and Kate Baumli and Shariq Iqbal and Colton Bishop and Rebecca Roelofs and Lei M Zhang and Kay McKinney and Disha Shrivastava and Cosmin Paduraru and George Tucker and Doina Precup and Feryal Behbahani and Aleksandra Faust},
booktitle={The Thirteenth International Conference on Learning Representations},
year={2025},
url={https://openreview.net/forum?id=CjwERcAU7w},
publisher={OpenReview.net}
}

@misc{kool2019,
title={Buy 4 {REINFORCE} Samples, Get a Baseline for Free!},
author={Wouter Kool and Herke van Hoof and Max Welling},
year={2019},
url={https://openreview.net/forum?id=r1lgTGL5DE},
note={ICLR 2019 Workshop drlStructPred}
}

@article{deepseek-r1,
  title={Deepseek-r1: Incentivizing reasoning capability in llms via reinforcement learning},
  author={Guo, Daya and Yang, Dejian and Zhang, Haowei and Song, Junxiao and Zhang, Ruoyu and Xu, Runxin and Zhu, Qihao and Ma, Shirong and Wang, Peiyi and Bi, Xiao and others},
  journal={arXiv preprint arXiv:2501.12948},
  year={2025}
}

@inproceedings{ouyang2022,
  title={Training language models to follow instructions with human feedback},
  author={Ouyang, Long and Wu, Jeffrey and Jiang, Xu and Almeida, Diogo and Wainwright, Carroll and Mishkin, Pamela and Zhang, Chong and Agarwal, Sandhini and Slama, Katarina and Ray, Alex and others},
  booktitle={Advances in neural information processing systems},
  volume={35},
  pages={27730--27744},
  year={2022},
  publisher={Curran Associates, Inc.}
}

@inproceedings{schulman2016high,
  title={High-dimensional continuous control using generalized advantage estimation},
  author={Schulman, John and Moritz, Philipp and Levine, Sergey and Jordan, Michael and Abbeel, Pieter},
  year={2016},
  booktitle={International Conference on Learning Representations (ICLR)},
  url={https://arxiv.org/abs/1506.02438},
  publisher={OpenReview.net}
}

@article{sutton1999,
  title={Policy gradient methods for reinforcement learning with function approximation},
  author={Sutton, Richard S and McAllester, David and Singh, Satinder and Mansour, Yishay},
  journal={Advances in neural information processing systems},
  volume={12},
  year={1999}
}

@article{survey1,
  title={A preliminary study of o1 in medicine: Are we closer to an ai doctor?},
  author={Xie, Yunfei and Wu, Juncheng and Tu, Haoqin and Yang, Siwei and Zhao, Bingchen and Zong, Yongshuo and Jin, Qiao and Xie, Cihang and Zhou, Yuyin},
  journal={arXiv preprint arXiv:2409.15277},
  year={2024}
}

@article{survey2,
  title={General scales unlock ai evaluation with explanatory and predictive power},
  author={Zhou, Lexin and Pacchiardi, Lorenzo and Mart{\'\i}nez-Plumed, Fernando and Collins, Katherine M and Moros-Daval, Yael and Zhang, Seraphina and Zhao, Qinlin and Huang, Yitian and Sun, Luning and Prunty, Jonathan E and others},
  journal={arXiv preprint arXiv:2503.06378},
  year={2025}
}

@article{survey3,
  title={Large Language Model Agents for Biomedicine: A Comprehensive Review of Methods, Evaluations, Challenges, and Future Directions},
  author={Xu, Xiaoran and Sankar, Ravi},
  journal={Information},
  volume={16},
  number={10},
  pages={894},
  year={2025},
  publisher={MDPI}
}

@article{ehrmind,
      title={Training LLMs for EHR-Based Reasoning Tasks via Reinforcement Learning}, 
      author={Jiacheng Lin and Zhenbang Wu and Jimeng Sun},
      year={2025},
      journal={arXiv preprint},
      doi={10.48550/arXiv.2505.24105}, 
}

@article{med-u1,
      title={Med-U1: Incentivizing Unified Medical Reasoning in LLMs via Large-scale Reinforcement Learning}, 
      author={Xiaotian Zhang and Yuan Wang and Zhaopeng Feng and Ruizhe Chen and Zhijie Zhou and Yan Zhang and Hongxia Xu and Jian Wu and Zuozhu Liu},
      year={2025},
      journal={arXiv preprint},
      doi={10.48550/arXiv.2506.12307}, 
}

@article{riskagent,
	author = {Liu, Fenglin and Wu, Jinge and Zhou, Hongjian and Gu, Xiao and Molaei, Soheila and Thakur, Anshul and Clifton, Lei and Wu, Honghan and Clifton, David A.},
	title = {RiskAgent: Autonomous Medical AI Copilot for Generalist Risk Prediction},
	year = {2025},
	doi = {10.1101/2025.04.03.25323489},
    journal={medRxiv preprint}
}

@article{oh2025,
  title={Rethinking Test-Time Scaling for Medical AI: Model and Task-Aware Strategies for LLMs and VLMs},
  author={Oh, Gyutaek and Kim, Seoyeon and Park, Sangjoon and Kim, Byung-Hoon},
  journal={arXiv preprint},
  year={2025},
  doi={10.48550/arXiv.2506.13102}
}

@inproceedings{score-to-steps,
    title = "From Scores to Steps: Diagnosing and Improving {LLM} Performance in Evidence-Based Medical Calculations",
    author = "Wang, Benlu  and  Xia, Iris  and Zhang, Yifan  and  Wang, Junda  and  Feiyun Ouyang  and  Han, Shuo  and  Cohan, Arman  and  Yu, Hong  and  Yao, Zonghai",
    booktitle = "Proceedings of the 2025 Conference on Empirical Methods in Natural Language Processing",
    month = "11",
    year = "2025",
    address = "Suzhou, China",
    publisher = "Association for Computational Linguistics",
    pages = "10820--10844",
    doi={10.18653/v1/2025.emnlp-main.548},
    ISBN = "979-8-89176-332-6"
}

@article{pg-williams,
  title={Simple statistical gradient-following algorithms for connectionist reinforcement learning},
  author={Williams, Ronald J},
  journal={Machine learning},
  volume={8},
  number={3},
  pages={229--256},
  year={1992},
  publisher={Springer}
}

@article{grpo,
  title={Deepseekmath: Pushing the limits of mathematical reasoning in open language models},
  author={Shao, Zhihong and Wang, Peiyi and Zhu, Qihao and Xu, Runxin and Song, Junxiao and Bi, Xiao and Zhang, Haowei and Zhang, Mingchuan and Li, YK and Wu, Yang and others},
  journal={arXiv preprint arXiv:2402.03300},
  year={2024}
}

@misc{mdcalc,
  title        = {Frequently Asked Questions},
  author       = {MDCalc},
  year         = {2024},
  howpublished = {\url{https://web.archive.org/web/20240405155011/https://www.mdcalc.com/faq}},
}

@article{goodell2025,
  title={Large language model agents can use tools to perform clinical calculations},
  author={Goodell, Alex J and Chu, Simon N and Rouholiman, Dara and Chu, Larry F},
  journal={npj Digit Med},
  volume={8},
  number={1},
  pages={163},
  year={2025},
  publisher={Nature Publishing Group UK London},
  doi={10.1038/s41746-025-01475-8}
}

@inproceedings{medcalc,
  title={Med{C}alc-{B}ench: Evaluating {L}arge {L}anguage {M}odels for {M}edical {C}alculations},
  author={Khandekar, Nikhil and Jin, Qiao and Xiong, Guangzhi and Dunn, Soren and Applebaum, Serina and Anwar, Zain and Sarfo-Gyamfi, Maame and Safranek, Conrad and Anwar, Abid and Zhang, Andrew and others},
  booktitle={Advances in Neural Information Processing Systems},
  volume={37},
  pages={84730--84745},
  year={2024},
  doi={10.52202/079017-2690},  url="https://proceedings.neurips.cc/paper_files/paper/2024/file/99e81750f3fdfcaf9613db2dbf4bd623-Paper-Datasets_and_Benchmarks_Track.pdf",
  note={(NeurIPS 2024 Datasets and Benchmark Track Oral)},
  publisher={Curran Associates, Inc.}
}

@article{medic-mohsen,
  title={Large language models for preventing medication direction errors in online pharmacies},
  author={Pais, Cristobal and Liu, Jianfeng and Voigt, Robert and Gupta, Vin and Wade, Elizabeth and Bayati, Mohsen},
  journal={Nat Med},
  volume={30},
  number={6},
  pages={1574--1582},
  year={2024},
  publisher={Nature Publishing Group US New York},
  doi={10.1038/s41591-024-02933-8}
}

\newpage
\begin{appendices}

\section{LLM Prompts}\label{app:prompts}

\begin{figure}[htbp]
\centering
\begin{tikzpicture}[
  font=\small,
  node distance=6mm,
  stage/.style={draw, rounded corners=2pt, fill=blue!6, align=left, inner sep=3pt, minimum width=38mm},
  arrow/.style={-stealth, thick, draw=black!60},
  every node/.style={align=left}
]

\node[draw, rounded corners=1pt, fill=gray!5, anchor=north west, inner sep=6pt, font=\ttfamily\scriptsize] (bg) at (0,0) {%
\begin{minipage}{0.97\textwidth}
Qwen3-8B System Prompt:\\[2pt] 
\colorbox{gray!25}{%
\begin{minipage}{0.99\textwidth}

You are a helpful medical assistant who gives truthful answers to the user’s questions.\\The user provides a patient note as well as a medical calculation question about the patient.\\You first conduct the reasoning process inside <think> and </think> tags. After you have made your best effort to answer the question, you provide the user with the final answer inside <answer> </answer> tags, i.e., <answer>numeric\_answer\_value</answer> or <answer>unknown</answer>. Importantly, inside the <answer> </answer> tags, you should provide either the numeric/date answer without units or additional textual explanation, or "unknown" for patient notes lacking necessary information. Note that "unknown" should be used sparingly, only when key information for answering the question is clearly missing from the patient note and cannot be reasonably inferred.\\

Example response 1:\\
<think>\\
The user wants to calculate the patient's BMI.\\
Formula: weight (kg) / (height (m))$^2$.\\
Extracted from note: Weight is "70 kg", Height is "175 cm".\\
Convert height: 175 cm = 1.75 m.\\
Calculation: 70 / (1.75 * 1.75) = 22.857.\\
</think>\\
<answer>22.857</answer>\\

Example response 2:\\
<think>\\
The user wants to calculate the patient's BMI.\\
Formula: weight (kg) / (height (m))$^2$.\\
Extracted from note: Height is "175 cm".\\
The patient's weight is not mentioned in the note. I cannot calculate the BMI.\\
</think>\\
<answer>unknown</answer>

\end{minipage}%
}\\[4pt]

Qwen3-8B User Prompt Template:\\[2pt] 
\colorbox{gray!25}{%
\begin{minipage}{0.99\textwidth}
\ttfamily\scriptsize

Patient Note: \{patient\_note\}\\

Question: \{question\}\\

Response:

\end{minipage}%
}
\end{minipage}%
};

\end{tikzpicture}
\caption{Prompts for the controlled RL experiment in \S\ref{sec:rl}, which are identically used across the two comparison groups. These prompts are fed as context into a Qwen3-8B model checkpoint that parametrizes the RL policy.}
\label{fig:qwen-prompts}
\end{figure}

\begin{figure}[htbp]
\centering
\begin{tikzpicture}[
  font=\small,
  node distance=6mm,
  stage/.style={draw, rounded corners=2pt, fill=blue!6, align=left, inner sep=3pt, minimum width=38mm},
  arrow/.style={-stealth, thick, draw=black!60},
  every node/.style={align=left}
]

\node[draw, rounded corners=1pt, fill=gray!5, anchor=north west, inner sep=6pt, font=\ttfamily\scriptsize] (bg) at (0,0) {%
\begin{minipage}{0.97\textwidth}
Phase 1 System Prompt:\\[2pt] 
\colorbox{gray!25}{%
\begin{minipage}{0.99\textwidth}
You will read a user-provided python dictionary object that contains an anonymized patient note ('Patient Note'), a clinical computation question ('Question'), a GPT-generated ground truth answer ('Ground Truth Answer'), as well as the relevant medical parameters that the GPT extracted from the patient note ('Relevant Entities') before it solves the computation question.\\

Your job, as a knowledgeable medical expert, is to fact-check whether the GPT has performed the computation question correctly in accordance with commonly accepted clinical standards. Review the information carefully, and perform independent calculations of the question if you can, and lastly give your best judgment of if you think the original GPT-generated ground truth answer is correct or not. It is acceptable for the GPT-generated answer to differ slightly from your judgment due to rounding errors, but the deviation should not be egregious as to have major difference in clinical implications. The question's underlying calculator might be sourced from a popular medical calculator website called MDCalc.com, and you are encouraged to use the web search tool to look up more specifically explanations for the clinical criteria and cross-check the GPT's.\\

Below, you will help me grade one such example, which I have randomly sampled from my database. Note that you do not need to be critical for the sake of it. Most importantly, you should remain objective and only flag cases that are truly problematic. After your reasoning and search procedures, you should end your response by boxing your final Yes/No judgment in <answer> </answer> tags, where Yes means you think the GPT-generated ground truth answer is correct, and No means you think it is incorrect.

\end{minipage}%
}\\[4pt]

Phase 1 User Prompt Example:\\[2pt] 
\colorbox{gray!25}{%
\begin{minipage}{0.99\textwidth}
\ttfamily\scriptsize
\{\\
\hspace*{1em}"Unique ID": 1,\\
\hspace*{1em}"Calculator ID": 2,\\
\hspace*{1em}"Calculator Name": "Creatinine Clearance (Cockcroft-Gault Equation)",\\
\hspace*{1em}"Category": "lab",\\
\hspace*{1em}"Output Type": "decimal",\\
\hspace*{1em}"Note ID": "pmc-7671985-1",\\
\hspace*{1em}"Note Type": "Extracted",\\
\hspace*{1em}"Patient Note": "An 87-year-old man was admitted to our hospital for anorexia for several days, high-grade fever from the previous day, and liver dysfunction. Of note, he had a history of hypertension, diabetes mellitus (DM), and angina. Physical examination findings included: clear consciousness; height, 163 cm; weight, 48 kg; blood pressure, 66/40 mmHg; heart rate, 75/min; respiratory rate, 22/min...[\textit{Omitted content was present in the actual API call}]...causes of death were thus determined as (1) DIC and (2) ALF associated with diffuse large B-cell lymphoma.",\\
\hspace*{1em}"Question": "What is the patient's Creatinine Clearance using the Cockroft-Gault Equation in terms of mL/min? You should use the patient's adjusted body weight in kg instead of the patient's actual body weight if the patient is overweight or obese based on their BMI. If the patient's BMI is normal, set their adjusted body weight to the minimum of the ideal body and actual weight. If the patient is underweight, please set their adjusted body weight to their actual body weight.",\\
\hspace*{1em}"Relevant Entities": "{'sex': 'Male', 'age': [87, 'years'], 'weight': [48.0, 'kg'], 'height': [163.0, 'cm'], 'creatinine': [1.4, 'mg/dL']}",\\
\hspace*{1em}"Ground Truth Answer": "25.238",\\
\hspace*{1em}"Lower Limit": "23.9761",\\
\hspace*{1em}"Upper Limit": "26.4999",\\
\hspace*{1em}"Ground Truth Explanation": "The formula for computing Cockcroft-Gault is given by CrCl = ((140 - age) * adjusted weight * (gender\_coefficient)) / (serum creatinine * 72), where the gender\_coefficient is 1 if male, and 0.85 if female. The serum creatinine concentration is in mg/dL. The patient's gender is male, which means that the gender coefficient is 1. The patient is 87 years old. The concentration of creatinine is 1.4 mg/dL. The formula for computing the patient's BMI is (weight)/(height * height),...[\textit{Omitted content was present in our API call}]...we take the patient's weight, 48.0 kg, as the patient's adjusted weight needed for the Cockroft-Gault Equation. Using the Cockcroft-Gault equation: CrCl = ((140 - age) * adjusted weight * gender\_coefficient) / (serum creatinine * 72). Plugging the patient's values gives us ((140 - 87) * 48.0 * 1) / (1.4 * 72) = 25.238 mL/min. Hence, the patient's creatinine clearance is 25.238 mL/min."\\
\}
\end{minipage}%
}
\end{minipage}%
};

\end{tikzpicture}
\caption{Prompts for the LLM audit agent executing the Phase 1 workflow (\S\ref{sec:phase1_audit}.)}\label{fig:phase-1-prompts}
\end{figure}

\begin{figure}[htbp]
\centering
\begin{tikzpicture}[
  font=\small,
  node distance=6mm,
  stage/.style={draw, rounded corners=2pt, fill=blue!6, align=left, inner sep=3pt, minimum width=38mm},
  arrow/.style={-stealth, thick, draw=black!60},
  every node/.style={align=left}
]

\node[draw, rounded corners=1pt, fill=gray!5, anchor=north west, inner sep=6pt, font=\ttfamily\scriptsize] (bg) at (0,0) {%
\begin{minipage}{0.97\textwidth}
Phase 2 System Prompt:\\[2pt] 
\colorbox{gray!25}{%
\begin{minipage}{0.99\textwidth}
You will read a user-provided JSON string that contains an anonymized patient note (``Patient Note'') and a clinical computation question (``Question''). Your job is to calculate the answer to the question based on the patient note. You have access to a Google Search tool and a code execution tool to perform calculations, if needed.\\

Your operational protocol is as follows:\\[2pt]
1.~\textbf{Adopt Expert Persona}: Assume the role of a diligent clinician. Your primary responsibility is accuracy and adherence to standard and up-to-date medical guidelines.\\[2pt]
2.~\textbf{Identify Appropriate Scoring Rule or Formula}: You will be given a patient note and a question that specifies a clinical calculation task (e.g., CHA2DS2-VASc score, MELD score, CURB-65, etc.). You should use Google Search to confirm the current, generally accepted formula, criteria, and/or scoring system for the specified clinical computation question. Prioritize reputable medical sources, such as MDCalc.com and guidelines published by health agencies or hospitals.\\[2pt]
3.~\textbf{Systematic Reasoning and Calculation}:\\
\hspace*{1em}$\bullet$~Carefully read the patient note and systematically extract every piece of information relevant to the calculation.\\
\hspace*{1em}$\bullet$~Explicitly map the extracted patient data to the components of the scoring system. For example, for a CHA2DS2-VASc score, list each component (Congestive heart failure, Hypertension, Age, etc.) and assign points based on the patient note.\\
\hspace*{1em}$\bullet$~If there are complex arithmetic operations that are not easily performed by hand, you can use the code execution tool to perform the calculations.\\
\hspace*{1em}$\bullet$~Show your step-by-step calculation. Sum the points or apply the formula as required.\\[2pt]
4.~\textbf{Handling Missing Information}:\\
\hspace*{1em}$\bullet$~If a piece of information required for the calculation is clearly absent from the note and if it cannot be reasonably inferred from the note, explicitly state what is missing in your reasoning, and give your final answer as ``unknown'' (without quotes).\\
\hspace*{1em}$\bullet$~Do not invent data or make assumptions that are not reasonably supported by the patient note.\\[2pt]
5.~\textbf{Final Answer Formatting (Strict)}:\\
\hspace*{1em}$\bullet$~After your detailed reasoning, you MUST enclose your final answer in \texttt{<answer>} \texttt{</answer>} tags.\\
\hspace*{1em}$\bullet$~The content inside the tags must ONLY be the final numeric value (e.g., 4, 12.5), date, or time. For numeric values, you do not need to round the answer.\\
\hspace*{1em}$\bullet$~As mentioned in \#4, if and only if the calculation is impossible due to missing data, the content inside the answer tags must be exactly ``unknown'' (without quotes).
\hspace*{1em}$\bullet$~DO NOT include any units, variable names, explanations, or any other text inside the \texttt{<answer>} \texttt{</answer>} tags.

\end{minipage}%
}\\[4pt]

Phase 2 User Prompt Example:\\[2pt] 
\colorbox{gray!25}{%
\begin{minipage}{0.99\textwidth}
\ttfamily\scriptsize

\{\\
\hspace*{1em}"Patient Note": "An 87-year-old man was admitted to our hospital for anorexia for several days, high-grade fever from the previous day, and liver dysfunction. Of note, he had a history of hypertension, diabetes mellitus (DM), and angina. Physical examination findings included: clear consciousness; height, 163 cm; weight, 48 kg; blood pressure, 66/40 mmHg; heart rate, 75/min; respiratory rate, 22/min...[\textit{Omitted content was present in the actual API call}]...causes of death were thus determined as (1) DIC and (2) ALF associated with diffuse large B-cell lymphoma.",\\
\hspace*{1em}"Question": "What is the patient's Creatinine Clearance using the Cockroft-Gault Equation in terms of mL/min? You should use the patient's adjusted body weight in kg instead of the patient's actual body weight if the patient is overweight or obese based on their BMI. If the patient's BMI is normal, set their adjusted body weight to the minimum of the ideal body and actual weight. If the patient is underweight, please set their adjusted body weight to their actual body weight.",\\
\}

\end{minipage}%
}
\end{minipage}%
};

\end{tikzpicture}
\caption{Prompts for the LLM relabeling agent executing the Phase 2 workflow (\S\ref{sec:phase2_relabel}). Note that unlike Fig. \ref{fig:phase-1-prompts}, the user prompt only includes $(C,q)$ to increase the independence of the relabeling process.}
\label{fig:phase-2-prompts}
\end{figure}

\section{Concurrent Work}\label{app:concur-work}
Wang et al.~\cite{score-to-steps} reexamined MedCalc-Bench by introducing an LLM-judged, stepwise evaluation pipeline to grade an LLM's intermediate calculation steps at test time, which can be thought of as providing process-level rewards besides a binary final answer correctness. Although they recommended to remove 10.3\% of the original test instances based on a clinician review, they otherwise treated the rest of the original MedCalc-Bench reference labels as valid. Our work is complementary yet distinct in our focus on developing a phased, physician-in-the-loop stewardship protocol that uses automated triage to amplify the impact of precious physician feedback, ultimately flagging 27.3\% of original test instances as likely label errors and independently recomputing their labels. The test instances we updated is more than 2.6 times as theirs. Furthermore, we show the downstream impact of the label noise by showing that \emph{training} a language model (using RL) on the revised versus original MedCalc-Bench reference labels would lead to a sizable distortion of perceived test accuracy, underscoring the importance of reference label choice.

Since November 2025, the MedCalc-Bench authors have been posting dataset updates, starting with v1.1 and v1.2,\footnote{Permanent URLs to \href{https://web.archive.org/web/20260121013009/https://huggingface.co/datasets/ncbi/MedCalc-Bench-v1.1/tree/3dde445d9aadb4f6f2af5d01f1400c0e4efc1a0a}{(v1.1)} and \href{https://huggingface.co/datasets/ncbi/MedCalc-Bench-v1.2/tree/915b1aad1fd2aa67790f369658cc365dfd440e90}{(v1.2)} of the MedCalc-Bench dataset, which were posted by a MedCalc-Bench author during November 2025 as two separate data repositories on Hugging Face.} on both GitHub and Hugging Face; each version release was presented as a replacement of the original v1.0 dataset published in NeurIPS conference. Then in December 2025, they started a new GitHub repository named ``MedCalc-Bench-Verified'', which describes itself as a replacement for v1.2,\footnote{\href{https://github.com/nikhilk7153/MedCalc-Bench-Verified}{https://github.com/nikhilk7153/MedCalc-Bench-Verified}. The new repository describes itself as a successor of the original (v1.0) repository at \href{https://github.com/ncbi-nlp/MedCalc-Bench}{https://github.com/ncbi-nlp/MedCalc-Bench}.} though the exact mechanism in which this new release was ``verified'' relative to the peer-reviewed NeurIPS v1.0 release remains unspecified. The latter repository's content was last revised on January 19, 2026, appearing to be unfinished as of our writing.

We view their ongoing revisions as beneficial and consistent with our thesis that LLM-assisted clinical benchmarks are living documents. The MedCalc-Bench authors do provide short versioned release notes (e.g., calculator bug fixes, replacement of some mismatched notes, etc.), but they are high-level and do not specify how reference-label curation, exclusion criteria, and dataset composition evolved across releases. These frequent updates/overwrites renders the ``the latest dataset snapshot'' a moving target, making it difficult for an external re-audit to yield conclusions that are scientifically generalizable as those from v1.0. In addition, downstream users do not uniformly migrate to the newest revision: in Appendix \S\ref{app:claude}, our independent reproduction study suggests that Anthropic’s reported ``Claude for Healthcare'' MedCalc-Bench accuracies are substantially closer to results obtained on v1.0 than on v1.2, suggesting that the original NeurIPS release remains a widely used baseline in practice. All audits and maintenance in this paper are conducted on their original, peer-reviewed NeurIPS release (v1.0),\footnote{For results reproducibility, our \href{https://github.com/junzeye/validate-medcalc-labels/tree/main/data/phase2}{GitHub repository} includes the MedCalc-Bench v1.0 \texttt{test} and \texttt{train} csv files used in our study.} the official version available when we conducted the dataset audit, label recomputations, and RL experiment in July-October 2025.
\section{Test-wide Error Estimation and Sources of Disagreement}\label{app:sec33-math}

In Phase 3 (\S\ref{sec:phase3-results}), we reported sMAPE values comparing the original ($\estimatorOrig$) and recomputed ($\estimatorNew$) labels against physician ground truth on a 50-instance subsample of the test set. This appendix asks: how does this comparison translate to the entire 1{,}047-instance test set? Direct measurement is infeasible because producing physician ground truth for the remaining instances would require an order of magnitude more clinician time than was available. We instead derive a conservative estimate from two ingredients.

First, our 50 physician-adjudicated instances were drawn from the 286 cases Phase 2 flagged as likely errors. We assume that the sMAPE values measured on this sample are representative of the full flagged stratum: i.e., that physician disagreement would scale similarly if more flagged cases were adjudicated. This is the only true assumption we make.

Second, on the 761 \emph{unflagged} test instances, $\estimatorOrig$ and $\estimatorNew$ are within $\pm 5\%$ of each other by construction. This is not an assumption but a fact: it follows directly from how Phase 2's triage rule defines ``unflagged.'' The implication is that on the unflagged stratum, the two estimators cannot meaningfully disagree on accuracy: if the original label is close to ground truth, the recomputed one must be too, and vice versa. We use this fact to bound the worst-case sMAPE on the unflagged stratum.

Combining these two ingredients yields a stratified test-wide estimate that conservatively favors the original benchmark. The remainder of this appendix formalizes this argument.

\paragraph{Setup and strata.}
On MedCalc's full test set ($n=1047$), Phase 2 (\S\ref{sec:phase2-results}) produces a high-confidence $\estimatorNew$ label for $887$ instances, in which $286$ are flagged as ``Likely Error'' ($220$ with $\textsf{rel.err}(\estimatorOrig,\estimatorNew)>0.05$ plus $66$ $\nullnotation$); the remaining $761=1047-286$ are unflagged. Phase 3 (\S\ref{sec:phase3-results}) then samples $50$ instances from the $286$ flagged instances for physician recomputation, reporting $\mathrm{sMAPE}(\estimatorOrig, y^\star)=72.7\%$ (95\% CI [47.9, 99.1]) vs $\mathrm{sMAPE}(\estimatorNew, y^\star)=20.1\%$ (95\% CI [8.0, 35.9]) against physician gold-standard labels ($y^\star$).

\paragraph{Worst-case assumption on unflagged majority.}
Adding robustness to our derivation, we adopt a conservative ``worst-case'' stance that significantly favors $\estimatorOrig$ by assuming it is perfectly accurate ($y^\star = \estimatorOrig$) for all 761 unflagged instances, an assumption unlikely to hold across the full set. Moreover, because $\mathrm{sMAPE}$ is computed only on pairs where both the prediction and $y^\star$ are numeric, any physician-labeled case with $y^\star=\nullnotation$ contributes no error. This effectively credits a numeric $\estimatorOrig$ prediction with (which always outputs a numeric value) $0\%$ error even when it hallucinates an answer. If such $\nullnotation$ vs. numeric mismatches were instead treated as large error (e.g. $100\%$), the resulting test-wide estimate would be even less favorable to $\estimatorOrig$ and would further widen the gap in favor of $\estimatorNew$.

\paragraph{Bounding $(\estimatorNew, y^\star)$'s sMAPE on unflagged instances.}
Phase 2 defines
\[
\textsf{rel.err}(a,b):=\frac{|a-b|}{\max(|a|,|b|)},
\;\;\;\text{and flags an instance if}\;\;\;
\textsf{rel.err}(a,b)>0.05.
\]
Recall also that $\mathrm{sMAPE}$ (before averaging over a batch):
\[
\mathrm{sMAPE}(a,b)=100\%\cdot \frac{2|a-b|}{|a|+|b|}.
\]
If $\textsf{rel.err}(a,b)\le \varepsilon$, and let $|b|=\max(|a|,|b|)$ without loss of generality, then $|a-b|\le \varepsilon |b|$ and $|a|\ge (1-\varepsilon)|b|$, so
\[
\mathrm{sMAPE}(a,b)\le 100\%\cdot \frac{2\varepsilon |b|}{(1-\varepsilon)|b|+|b|}
=100\%\cdot \frac{2\varepsilon}{2-\varepsilon}.
\]
Since $\varepsilon \leq 0.05$ for all unflagged instances by definition, $\mathrm{sMAPE}(a,b)\le 100\%\cdot \frac{0.10}{1.95}\leq 5.13\%$.

\paragraph{Test-wide sMAPE estimation.}
Use $s_{\cdot}$ to abbreviate $\mathrm{sMAPE}(\cdot,y^\star)$, with the former's subscript ``$_{\cdot}$'' specifying the estimator we evaluate. Treat the Phase 3 physician-adjudicated samples' sMAPE values as a proxy for the flagged stratum, i.e., $s_{\text{orig,flag}}=72.7\%$ and $s_{\text{new,flag}}=20.1\%$. For the unflagged majority, under $y^\star=\estimatorOrig$, take $s_{\text{orig,unflag}}=0$ and $s_{\text{new,unflag}}\le 5.13\%$ (numeric-case bound above). Then
\[
\bar s_{\text{orig}}\approx \frac{286}{1047}\cdot 72.7 + \frac{761}{1047}\cdot 0
\geq 19.8\%,
\qquad
\bar s_{\text{new}}\approx \frac{286}{1047}\cdot 20.1 + \frac{761}{1047}\cdot 5.13
\leq 9.3\%.
\]
We emphasize this is an \emph{estimated} test-wide summary (Table~1 sMAPE is computed on the numeric physician-labeled subset), not a directly observed average over all $1{,}047$ instances.

\paragraph{Scaled Confidence Intervals for Test-wide sMAPE}
Since the unflagged stratum ($n_{\text{unflag}}=761$) is treated as having fixed constant error values ($K$) under our counterfactual assumptions, the variance in the test-wide estimates $\bar{s}$ arises solely from the sampling uncertainty in the flagged stratum ($n_{\text{flag}}=286$). We therefore derive the test-wide confidence intervals by linearly scaling the bootstrap confidence intervals from Figure \ref{fig:phase3-result}.

The test-wide estimator $\bar{s}$ is a linear combination:
\begin{equation}
    \bar{s} = \frac{n_{\text{flag}}}{N} \cdot s_{\text{flag}} + \frac{n_{\text{unflag}}}{N} \cdot K
\end{equation}
where $N=1{,}047$, $n_{\text{flag}}=286$, and $n_{\text{unflag}}=761$. If $[L_{\text{flag}}, U_{\text{flag}}]$ is the 95\% confidence interval for the physician-adjudicated sample $s_{\text{flag}}$, the derived interval for the test-wide mixture is:
\[
\left[ \frac{286}{1047} L_{\text{flag}} + \frac{761}{1047} K, \quad \frac{286}{1047} U_{\text{flag}} + \frac{761}{1047} K \right]
\]

For $\bar{s}_{\text{orig}}$, we use $K=0$ and the Table 1 interval $[47.9, 99.1]$:
\begin{align*}
    CI_{\text{orig}} &= \left[ \frac{286}{1047} (47.9), \quad \frac{286}{1047} (99.1) \right] \\
    &\approx [13.1, 27.1]
\end{align*}

For $\bar{s}_{\text{new}}$, we use the conservative bound $K=5.13$ and the Table 1 interval $[8.0, 35.9]$. Note that the scaled constant term contributes $\frac{761}{1047} \cdot 5.13 \approx 3.73\%$:
\begin{align*}
    CI_{\text{new}} &= \left[ \frac{286}{1047} (8.0) + 3.73, \quad \frac{286}{1047} (35.9) + 3.73 \right] \\
    &\approx [5.9, 13.5]
\end{align*}

Even under this ultra-conservative scenario where $\estimatorOrig$ is assumed perfect on all the unflagged test instances, which is extremely unlikely to hold, their confidence intervals are substantially separated: the \emph{upper} bound of the new labels' error (13.5\%) only marginally crosses the \emph{lower} bound of the original labels' error (13.1\%), indicating that $\estimatorNew$'s clinical alignment advantage is statistically robust to sampling uncertainty and conservative imputation assumptions.

\subsection{Qualitative Sources of Disagreement}

Beyond the quantitative argument above, the physicians' free-text comments collected during Phase 3 surface recurring \emph{qualitative} reasons why a single ``ground truth'' label may not exist for many instances. We summarize these patterns below.
Retrospective comparison against single-blind physician recomputations suggests that a substantial share of label disagreement is driven by \emph{context underspecification}: many original $(C,q)$ instances do not contain enough contextual qualifiers to define a unique clinical ``ground truth'' score. Two patterns recur. First, the benchmark rarely specifies a scoring \emph{timepoint}. Many notes describe multi-visit trajectories over weeks or months, but the question does not clarify whether the score should be computed at presentation, peak severity, discharge, or another reference time (e.g. rows 138, 334, 341, 509, 569, 739, 761, 785). Second, scoring rules can vary across time and jurisdiction, so the ``official'' aggregation rule is not uniquely describable by a static benchmark label. For example, Morphine Milligram Equivalent (MME) conversions for methadone depend on which guideline is assumed: pre-2023 CDC guidance used a tiered rule, i.e. a multiplier of 12 for methadone doses $>60$ mg/day \cite{cdc_opioids_2016}, whereas the current CDC guideline adopts a flatter multiplier (4.7) while cautioning that conversion becomes unreliable at higher doses \cite{dowell2022cdc}; still, today's evidence suggest that some state-level guidelines reflect the older CDC schedule \cite{maryland_mme, utah_mme}. Under such ambiguity, multiple defensible answers exist and a single numeric gold label does not cleanly measure an LLM's generalizable clinical capability.
\section{Technical Background: From Score Computation to Model Training}\label{sec:medical-score}

This appendix collects the formal definitions and derivations that underpin the main text. It is organized in four parts. \S\ref{subsec:math} formalizes medical score computation as a two-stage pipeline (feature extraction followed by rule-based aggregation), identifying the points at which labeling errors can enter; Table~\ref{tab:formalism-lace-mapping} grounds this formalism in a concrete LACE score example. \S\ref{sec:current-standard} describes MedCalc-Bench's evaluation protocol and summarizes prior reported results. \S\ref{sec:estimation-lens} discusses what it means to treat a benchmark label as a point estimate of a latent ground truth, and the implications for inter-rater variability. Finally, \S\ref{sec:model-alignment-mdp} provides a self-contained derivation of the GRPO algorithm used in our RL experiment, adapted to the on-policy, trajectory-level reward setting relevant to medical score computation.

\begin{table}[htbp]
\centering
\caption{Mapping Formalism to a Concrete Example: The LACE Score \cite{lace}}
\label{tab:formalism-lace-mapping}
\renewcommand{\arraystretch}{1.3}
\setlength{\tabcolsep}{8pt}
\footnotesize
\begin{tabular}{>{\centering\arraybackslash}p{2.5cm}p{4cm}p{5.5cm}}
\toprule
\rowcolor{headerblue}
\textcolor{white}{\textbf{Mathematical}} & \textcolor{white}{\textbf{Formal}} & \textcolor{white}{\textbf{LACE Score for Readmission}} \\
\textcolor{white}{\textbf{Symbol}} & \textcolor{white}{\textbf{Definition}} & \textcolor{white}{\textbf{Example}} \\
\midrule
\rowcolor{lightgray}
$C$ & Patient context (EHR notes and lab data) & \textit{``72M admitted to ED for CHF exacerbation. Hospital stay was 4 days. Hx of diabetes (CCI=2). Visited the ED twice in last 5 months.''} \\
$q$ & Medical score to compute & \textit{``What is the patient's LACE score?''} \\
\rowcolor{lightgray}
$(q_i)_{i=1}^{m}$ & Sub-questions for each feature & 
\begin{tabular}[t]{@{}l@{}}
$q_1$: Patient's \textbf{L}ength of stay?\\
$q_2$: \textbf{A}cuity of admission?\\
$q_3$: \textbf{C}harlson Comorbidity Index?\\
$q_4$: \# \textbf{E}D visits in last 6 months?
\end{tabular} \\
\rowcolor{lightblue}
$f_\theta(C, q_i)$ & \textbf{Stage 1:} Feature extractor (language processing) & 
\begin{tabular}[t]{@{}l@{}}
Clinician/LLM reads $C$ to find\\
answers to each $q_i$
\end{tabular} \\
\rowcolor{lightgray}
$\bm{x} \triangleq (x_i)_{i=1}^{m}$ & Clinical features vector, where $m = 4$ for LACE & 
\begin{tabular}[t]{@{}l@{}}
$x_1 = $ 4 days ($\in \mathbb{R}$)\\
$x_2 = $ Emergent ($\in \mathcal{C}$)\\
$x_3 = $ CCI score of 2 ($\in \mathbb{R}$)\\
$x_4 = $ 2 visits ($\in \mathbb{R}$)
\end{tabular} \\
\rowcolor{lightblue}
$g_\phi(x_1,...,x_m)$ & \textbf{Stage 2:} Aggregation function (rule-based reduction) & 
\begin{tabular}[t]{@{}l@{}}
Rule $\phi$: Map to points then sum\\
4 days $\rightarrow$ 4pts, Emergent $\rightarrow$ 3pts\\
CCI 2 $\rightarrow$ 2pts, 2 visits $\rightarrow$ 2pts\\
$g_\phi = 4+3+2+2 = \mathbf{11}$
\end{tabular} \\
\rowcolor{lightgray}
$y$ & Final computed score & \textbf{11} \\
$G_{\phi,\theta}[C,q]$ & End-to-end pipeline composition & Raw EHR note $\rightarrow$ Extract 4 features $\rightarrow$ Apply LACE rules $\rightarrow$ \textbf{Score = 11} \\
\bottomrule
\end{tabular}
\end{table}

\subsection{Mathematical formalism}\label{subsec:math}
A typical workflow of computing a medical score can be decomposed into two stages: (1) evaluating a fixed set of score-specific clinical features (a.k.a. clinical criteria) based on the patient's medical history, then (2) formulaically aggregating the feature values into a scalar score, either by hand or a custom calculator app like \href{http://www.MDCalc.com}{MDCalc}. 
In symbols, we may denote the patient context (natural language texts interleaved with tabular lab data) by $C$ and a description of the medical score to compute by $q$. The feature vector typically used to answer $q$ can be denoted by $\bm{x} \triangleq (x_i)_{i=1}^{m}$ where $m \in \mathbb{N}$ is uniquely determined by $q$. $(x_i)_{i=1}^{m}$ can be thought of as answers to $(q_i)_{i=1}^{m}$ where $q_i$ is a sub-question of $q$, described in natural language, that a clinician or LLM needs to answer. For each $i$, the per-feature domain $\mathcal{X}_i$ contains the following:
\begin{enumerate}
    \item real numbers, e.g. blood pressure, Likert scale response;
    \item categorical values, ranging from True/False to symptom groups; we call this a set $\cC$;
    \item a fallback undefined token denoted by $\nullnotation$, indicating unextractability of a feature.
\end{enumerate}
In sum, $x_i \in \mathcal{X}_i := \mathbb{R} \cup \cC \cup \{\nullnotation\}$. Let $\bm{\mathcal{X}} \triangleq \prod_{i=1}^m \mathcal{X}_{i}$ denote the feature space, a Cartesian product of individual per-feature domains. Lastly, for each $(C,q)$ let $y$ be \emph{any} answer predicted by an algorithm, and $y^*$ be the (latent) clinically correct ground truth answer; $y^*$ is a scalar if the question $q$ is answerable given $C$, and $\nullnotation$ if $C$ lacks crucial information or is not clinically suited for answering $q$. Perfectly observing $y^*$ is costly, so an estimator $\estimator$ ($\estimatorOrig$ for MedCalc-Bench, $\estimatorNew$ for ours) is used.

\paragraph{Stage (1): Extracting Features.}
With the above notations, and let $\mathcal{T}$ be the set of all displayable computer texts. The first stage of the workflow is abstracted as applying a highly flexible language processing function $f_{\theta}: \mathcal{T} \rightarrow \bigcup_{i} \mathcal{X}_i$ for each one of $x_i = f_{\theta}(C,q_i)$ for $1\leq i\leq m$, i.e., $m$ times applying $f_\theta$. The language processor $f_{\theta}$ is traditionally instantiated by a human healthcare provider who reads patient EHR notes to determine the $q_i$'s, but nowadays the rise of generative AI makes it possible to augment human-only processing with LLM auto-completions.

\paragraph{Stage (2): Aggregating Features.}
Given the feature vector \( \bm{x}\) extracted in Stage (1), the second stage applies a rule-based, deterministic reduction operator $g_\phi$ (Of course, $g_\phi$ is specific to each score question $q$, but for notational simplicity, we do not include $q$ in the notation) that maps the feature tuple $(x_i)_{i=1}^m$ to a scalar output $y \in \mathcal{Y}$. The parameter $\phi$ represents the score-specific rule, e.g. arithmetic formula, look-up table, or piecewise-defined condition, that defines how the features jointly contribute to the final score. Formally:
\[
g_\phi : \bm{\mathcal{X}} \mapsto \mathcal{Y},\quad y=g_\phi(\bm{x})=g_\phi(x_1,...,x_m)
\]
where \( \mathcal{Y} \subseteq \mathbb{R} \cup \{\nullnotation\}\) is the space of possible score values. If at least one feature in $\bm{x}$ is out-of-range or $\nullnotation$, the final score is not computable and $\nullnotation$ is returned.

Putting the two stages together, we can visualize the overall workflow of computing a medical score as a compositional pipeline:
\[
(C,q) \xrightarrow[\text{extracts}]{f_\theta}
(x_1, x_2, \ldots, x_m)
\xrightarrow[\text{aggregates}]{g_\phi}
y.
\]
The end-to-end pipeline can be abstracted by a composed function $G_{\phi,\theta}[C,q] : \mathcal{T} \mapsto \mathcal{Y}$, specifically 
\[G_{\phi,\theta}[C,q] \triangleq g_\phi(f_\theta(C, q_1), \ldots, f_\theta(C, q_m)).\]

\paragraph{Insights from the Formalism}
The above formalism abstracts the \emph{sequential modularity} of a typical medical computing workflow and the \emph{graphical structure} of information flows: the first stage ($f_\theta$) interprets and extracts structured intermediate variables from unstructured data, while the second stage ($g_\phi$) performs a rule-based routine to aggregate the intermediate bits into a clinically verifiable and actionable scalar output. Of course, not all end-to-end workflows follow this specific pipeline structure; a system designer may as well eliminate $g_\phi$ and use a single call of $f_{\theta}$ to directly output the scalar output. That said, \cite{medcalc} employed the former, structured data generation approach, so understanding this allows us to attribute any deviation in the final output $y$ to a parameter extraction error in $f_\theta$, or a formulaic mistake in $g_\phi$, or both.

\subsection{Current task evaluation standard}\label{sec:current-standard}
To our best knowledge, the only large-scale public benchmark dataset for evaluating an agentic system's end-to-end score computing capability is MedCalc-Bench, created and released by researchers at NIH/NLM and five US medical schools \cite{medcalc}. Their benchmark task asks an LLM model to read a patient note and answer a medical score computation question, and the model's final answer $y$ is graded correct if it is within a $\pm 5\%$ margin of a gold label value $\estimatorOrig$ that the benchmark authors synthesized; and in cases where $y$ should be a datetime, exact string match is required. A model's performance is quantified as the average accuracy on the benchmark's test data set, which are 1047 instances manually reviewed by those authors.

MedCalc-Bench authors curated the benchmark data and labels in the following manner: 
\begin{enumerate}
    \item Patient notes are real, deidentified patient cases they scraped from journal-published medical papers that are archived in the \href{https://pmc.ncbi.nlm.nih.gov/}{PubMed Central} database. There are altogether 10,053 + 1,047 = 11,100 notes in the train and test sets. We note that each of these patient notes can be seen as a unique context $C$ in the notation of \S\ref{subsec:math}. 
    \item The medical score questions ($q$'s), e.g. ``What is the patient's Glasgow Coma Score?'', are sourced as the 55 unique calculators listed as ``popular'' on \href{https://www.mdcalc.com}{MDCalc.com}, a web app widely used by U.S. physicians. One can think of each train or test instance in the MedCalc-Bench dataset as uniquely identified by a pair $(C,q)$. Many instances may share the same score question $q$, though their patient context $C$ will differ.
    \item $\estimatorOrig \in \mathcal{Y}$, MedCalc-Bench's ground truth label of each $(C,q)$, was synthesized in a procedure analogous to the compositional pipeline we presented in \ref{subsec:math}. They used \texttt{GPT-4} as $f_\theta$ and themselves hard-coded the logic of $g_\phi$ of each unique $q$ \href{https://github.com/ncbi-nlp/MedCalc-Bench/tree/main/calculator_implementations}{individually as 55 Python scripts}.
\end{enumerate}

\begin{table}[htbp]
\centering
\footnotesize
\caption{Some Reported MedCalc-Bench results.}
\label{tab:medcalc_sota}
\setlength{\tabcolsep}{6pt}
\begin{tabular}{l l l c}
\toprule
\textbf{Paper} & \textbf{Base Model} & \textbf{Setup} & \textbf{Acc.} \\
\midrule
Khandekar et al.~\cite{medcalc}   & GPT-4         & one-shot                 & 50.9\\
Oh et al.~\cite{oh2025}           & Llama3-70B    & 4-bit quantized          & 40.0            \\
Bedi et al.~\cite{medhelm}        & o3-mini         & zero-shot                 & 34.0\\
Wang et al.~\cite{score-to-steps} & GPT-4o        & RAG + code               & 59.6\textsuperscript{*} \\
Liu et al.~\cite{riskagent}       & GPT-4o        & multi-agent + tool use   & 67.7            \\
Zhang et al.~\cite{med-u1}        & Qwen2.5-7B    & RLVR                     & 57.6            \\
Lin et al.~\cite{ehrmind}         & DeepSeek-R1   & zero-shot                & 48.1            \\
Lin et al.~\cite{ehrmind}         & Llama3-3B     & SFT + RLVR               & 52.0            \\
\bottomrule
\end{tabular}

\vspace{2pt}
\raggedright\footnotesize ``Acc.'' is overall accuracy on the test set (\%) of MedCalc-Bench. ``Setup'' refers to each study's main adaptation technique of the base model. All studies reported using Chain-of-Thought (reasoning mode) in their LLM inference. \textsuperscript{*}This reported result was evaluated on a subset of the original MedCalc test set ($940/1047$).
\end{table}

Since June 2024, multiple academic studies have used MedCalc-Bench as an evaluation benchmark or finetuned open-weight models on its train set. Table \ref{tab:medcalc_sota} shows a partial anthology of some test accuracies recently reported. Several survey papers on LLM and healthcare, including \cite{survey1, survey2, survey3}, cited this benchmark. In addition, MedCalc-Bench is also highlighted as a representative quantitative clinical benchmark in the test suite of MedHELM \cite{medhelm}, a comprehensive framework for evaluating medical LLMs, recently published on \emph{Nature Medicine}. These references indicate that MedCalc-Bench is an influential benchmark in the field. As all these studies were posted online before October 2025, i.e. before MedCalc-Bench authors had publicly recommended any changes to their dataset version in November 2025, it is fairly certain that their evaluations were based on the original NeurIPS version, i.e. (v1.0).

\subsection{A statistical estimation lens of ground truth assurance}\label{sec:estimation-lens}
If ground truth $y^*$ were to be thought of as a point value and $\estimator$ be a benchmark's estimate of it, the alignment objective could be loosely interpreted as minimizing the estimator bias 
$B(\estimator):= \mathbb{E}_{\estimator\sim D(\theta, \phi)}[\estimator] - y^*$,
where $D(\theta, \phi)$ emphasizes the sampling distribution's dependence on both on the LLM weights $\theta$ and the rule-based aggregation logic $\phi$ stipulated by the human researcher. If there is high heterogeneity in how human clinicians themselves interpret a clinical criterion (low inter-rater reliability), $y^*$ might be better modeled as a distribution; and the alignment objective be approximated as minimizing some statistical distance, e.g. KL divergence or Wasserstein distance, between $y^*$ and $\estimator$. In this paper, we adopt the former, point-estimate view of $y^*$ with a $\pm5\%$ margin for real-valued outputs, reflecting the evaluation guideline given by MedCalc-Bench. That said, we want to acknowledge the possible suitability of the second approach for certain medical score questions.

\subsection{Model alignment as policy optimization}\label{sec:model-alignment-mdp}

We adopt a standard sequential decision-making formalism for model alignment, following the RL literature on policy-gradient methods and recent work on LLM post-training.\footnote{See, e.g. \cite{pg-williams, sutton1999, schulman2016high, kool2019, ouyang2022, ahmadian2024, deepseek-r1, kumar25}.} In this view, the interaction between an LLM agent and its digital environment is a Markov decision process (MDP) 
$(\mathcal{S}, \mathcal{A}, R, P, \mu)$: at each discrete time $t = 1,\dots,T$ the agent observes a state $s_t \in \mathcal{S}$, samples an action $a_t \in \mathcal{A}$ from a stochastic policy $\pi_\theta(a_t \mid s_t)$, receives a reward $r_t = R(s_t,a_t,s_{t+1})$, and the environment transitions to $s_{t+1} \sim P(\cdot \mid s_t,a_t)$. A trajectory (i.e. a token sequence for an LLM) is
\[
\tau \;=\; (s_1,a_1,s_2,a_2,\dots,s_{T},a_{T},s_{T+1}),
\]
with initial state $s_1 \sim \mu$. The policy objective is to maximize the expected return
\begin{equation}
J(\theta)
\;=\;\mathbb{E}_{\tau \sim p_\theta(\tau)}[R(\tau)]
\;=\;\int p_\theta(\tau) R(\tau)\, d\tau ,
\quad
R(\tau) \;=\; \sum_{t=1}^{T} r_t,
\label{eq:rl-objective}
\end{equation}
where $p_\theta(\tau)$ is the trajectory distribution induced by $\pi_\theta$ and the MDP dynamics.

\paragraph{Instantiating the MDP for MedCalc-Bench.}
For an LLM, the state $s_t$ is the current token sequence cached in the GPU memory (prompt, EHR snippet, and previously generated tokens), the action $a_t$ is the next token, and the transition is deterministic: $s_{t+1} = (s_t,a_t)$. The initial state $s_1 \sim \mu$ corresponds to a prompt i.e. $(C,q)$ randomly sampled from the set of MedCalc-Bench instances. In classical MDP formulations, returns are often written as a sum of stepwise rewards like the second equation in (\ref{eq:rl-objective}). However, in LLM post-training a simpler, \emph{outcome-based} interpretation is more often assumed. Given an input $(C,q):=x$ and a final answer $y(\tau)$ parsed from a completed text trajectory $\tau$, we set $r_t=0$ for all $t<T$ such that the trajectory-level reward becomes
\begin{equation}
R(\tau)\;=\;r_T
\;:=\;
\underbrace{\lambda_{\mathrm{f}}\,\cdot
\mathbf{1}\{\,\hat{y}(\tau)\text{ is parsable}\,\}}_{\text{format reward}}
\;+\;
(1-\lambda_{\mathrm{f}})\,\cdot \underbrace{\mathbf{1}\{\, \operatorname{rel\_err}(\hat y(\tau),y^*) \leq \epsilon \,\}}_{\text{answer reward}}
\label{eq:reward-function}
\end{equation}
that combines correctness of the computed score and adherence to the required output format for regex parsability, evaluated against a gold label $y^*$ (surrogated by either $\estimatorOrig$ or $\estimatorNew$). In this abstraction, varying $y^*$'s value corresponds exactly to changing the reward function $R$ in the MDP.

\paragraph{Policy gradient and REINFORCE.}
To optimize \eqref{eq:rl-objective}, we differentiate under the integral and apply the log-derivative identity $\nabla_\theta p_\theta(x) = p_\theta(x)\nabla_\theta \log p_\theta(x)$:
\begin{align}
\nabla_\theta J(\theta)\;=\;\int \nabla_\theta p_\theta(\tau)\, R(\tau)\, d\tau \nonumber\;=\;\mathbb{E}_{\tau \sim p_\theta(\tau)}\!\left[ R(\tau)\, \nabla_\theta \log p_\theta(\tau) \right].
\label{eq:policy-gradient}
\end{align}
Because $p_\theta(\tau)$ factors as
\[
p_\theta(\tau)
= \mu(s_1)\prod_{t=1}^{T} \pi_\theta(a_t \mid s_t)\, P(s_{t+1} \mid s_t,a_t),
\]
we obtain the vanilla REINFORCE policy gradient
\begin{equation}
\nabla_\theta J(\theta)
=
\mathbb{E}_{\tau \sim p_\theta(\tau)}
\left[
R(\tau)
\sum_{t=1}^{T} \nabla_\theta \log \pi_\theta(a_t \mid s_t)
\right].
\label{eq:reinforce}
\end{equation}
\newline
A standard variance-reduction trick is to subtract a zero-mean baseline $b$ that does not depend on the realized value of $\tau$:
\begin{equation}
\nabla_\theta J(\theta)
=
\mathbb{E}_{\tau \sim p_\theta(\tau)}
\left[
\big(R(\tau) - b\big)
\sum_{t=1}^{T} \nabla_\theta \log \pi_\theta(a_t \mid s_t)
\right].
\label{eq:reinforce-baseline}
\end{equation}
This is the \textit{REINFORCE with baseline} estimator in the trajectory-level reward setting we use: every token on a trajectory shares the same scalar credit signal $R(\tau) - b$.

\paragraph{Group-relative policy optimization (GRPO).}
GRPO specializes \eqref{eq:reinforce-baseline} by using a \emph{group-relative} baseline at the level of each initial state $s_1$ (think a $(C,q)$ tuple in MedCalc-Bench). Fix a group size $G$ (e.g. $8$), then for every $s_1$ we sample $G$ trajectories
$\{\tau^{(i)}\}_{i=1}^{G}$ under the current policy $\pi_{\theta}$, compute their trajectory-level rewards $R^{(i)} = R(\tau^{(i)})$, form the group average $\bar R$ and define advantages $A^{(i)}$
\begin{equation}
\bar R \;=\; \frac{1}{G}\sum_{i=1}^{G} R^{(i)},\quad A^{(i)} \;=\; R^{(i)} - \bar R ,
\label{eq:grpo-advantage}
\end{equation}
so that trajectory samples with above-average reward receive positive credit and those with below-average reward receive negative credit, finally back-propagated into LLM weight updates.

In our simplified on-policy implementation, the GRPO loss for updating $\theta$ is
\begin{equation}
L_{\mathrm{GRPO}}(\theta)
=
-\,\mathbb{E}_{\{\tau^{(i)}\}_{i=1}^{G} \sim p_\theta(\cdot \mid s_1)}
\left[
\sum_{i=1}^{G}
A^{(i)}
\sum_{t=1}^{T^{(i)}}
\log \pi_\theta\!\big(a_t^{(i)} \mid s_t^{(i)}\big)
\right]
\;+\;
\beta\,\cdot {\mathrm{KL}}\!\left(\pi_\theta \,\middle\|\, \pi_{\mathrm{ref}}\right),
\label{eq:grpo-loss}
\end{equation}
where $T^{(i)}$ is the length of trajectory $\tau^{(i)}$, $\pi_{\mathrm{ref}}$ is a fixed reference policy, and $\beta>0$ controls the strength of KL regularization. Minimizing \eqref{eq:grpo-loss} corresponds to a REINFORCE-style policy update that (i) increases the log-probability of completions whose reward is higher than their group mean and (ii) keeps the updated policy close to $\pi_{\mathrm{ref}}$ to avoid catastrophic forgetting.

In summary, under this MDP and GRPO formalism, surrogating the reference label $y^\star$ by $\estimatorOrig$ or $\estimatorNew$ shapes the reward assignment $R(\tau)$ in \eqref{eq:reward-function}, and therefore the advantages $A^{(i)}$ in \eqref{eq:grpo-advantage}. The rest of the policy-optimization pipeline is held fixed, so any difference in the fine-tuned $\pi_\theta$'s test accuracy can be attributed to the choice of labels used to define the alignment objective.

\section{Controlled RL Experiment Details}\label{app:RL}

\paragraph{Goal and overall design.}
Our goal is to isolate the effect of label choice on an LLM’s alignment when RL is used. We run two GRPO training experiments on the same base model that are identical in all respects except for the labels used to define the training reward. In the \emph{original-label} arm, training rewards (as assigned by \eqref{eq:reward-function}) are computed against MedCalc-Bench’s original labels $\estimatorOrig$; in the \emph{new-label} arm, training rewards are computed against our recomputed labels $\estimatorNew$.

\paragraph{Evaluation strategy.}
We evaluate the two trained models along three dimensions, each reported in a different section:
\begin{enumerate}
    \item \textbf{Primary evaluation against physician ground truth ($y^\star$, $n=50$).} Both models are evaluated on the 50 physician-labeled test instances from Phase 3, where we have direct access to clinician-computed labels. This is our primary measure of whether label quality during training translates into clinically meaningful differences. The main result is reported in the Results section (\S\ref{results:RL}).
    \item \textbf{Full training dynamics on the 887-instance test set (graded against $\estimatorNew$).} We track accuracy over the course of training on the larger held-out set, using $\estimatorNew$ as a proxy for ground truth. This provides a higher-powered view of the performance gap. Details are in Appendix~\S\ref{app:RL-887}; a summary appears in \S\ref{results:RL}.
    \item \textbf{Generalization to additional medical evaluation sets.} We test both checkpoints on MedQA and two recent revisions of MedCalc-Bench to assess whether the training-label effect transfers to related medical tasks. Details are in Appendix~\S\ref{app:RL-ood}; a summary appears in \S\ref{results:RL}.
\end{enumerate}

\paragraph{Dataset and label conditions.}
Both RL runs are trained on the same subset of $4{,}593$ MedCalc-Bench train instances for which Phase~2 produced high-confidence $\estimatorNew$ labels. The held-out test set consists of $887$ instances whose $\estimatorNew$ labels were also obtained from Phase~2 and partially audited by physicians. In both arms, the policy LLM sees exactly the same collection of prompts and calculator types; the only difference is whether the training reward’s $y^*$ is instantiated using $\estimatorOrig$ or $\estimatorNew$. The underlying system and user prompts, including chain-of-thought and answer-format instructions, are held fixed across runs; we provide the full prompt template in Appendix~\S\ref{app:prompts}.

\paragraph{Policy initialization and interaction environment.}
In both arms, the trainable policy $\pi_\theta$ and the frozen reference policy $\pi_{\text{ref}}$ are initialized from the same Qwen3-8B base model \cite{qwen3}, chosen as a widely used open model for academic post-training. $\pi_{\text{ref}}$, being used for KL regularization, remains fixed throughout training, while $\pi_\theta$ is updated iteratively by gradient descent to minimize the GRPO loss in Eq.~\ref{eq:grpo-loss}. We use the same updated system prompt (which allows for abstention) and require chain-of-thought style completions with a final scalar answer token; no external tools are used during RL training, and the interaction is purely text-based.

\paragraph{Key training hyperparameters.}
We instantiate the policy objective from \S\ref{sec:model-alignment-mdp} using the trajectory-level reward $r(\tau)$ in Eq.~\ref{eq:reward-function} with $\lambda_f=0.1$. For the comparison experiment in \S\ref{sec:rl}, both runs share the same optimizer and hyperparameters: Adam with learning rate $\text{lr} = 10^{-5}$, KL regularization coefficient $\beta = 10^{-3}$, and no reward standardization or length normalization to ensure advantage estimation is unbiased \cite{drgrpo}. Each GRPO minibatch consists of $256$ distinct $(C,q)$ pairs from the MedCalc training set. We maintain the same random seed so that the minibatch partitions and ordering are identical between the two runs. We also conducted a hyperparameter sensitivity analysis over the learning rates $\{10^{-4}, 10^{-5}, 10^{-6}\}$ and KL coefficient $\{10^{-2}, 10^{-3}, 10^{-4}\}$.

\paragraph{Sampling and evaluation protocol.}
During training, all trajectories are generated from $\pi_\theta$ with a temperature of $1.0$, top-$p = 1.0$, and a maximum output length of $1600$ tokens. Moreover, the GRPO group size $G = 8$, i.e. $8$ trajectories are sampled i.i.d. from $\pi_\theta$ for each $(C,q)$ pair in a minibatch. After training, we evaluate the two policies on the $887$-instance held-out test set, with ths same inference configuraiton as training rollouts. The evaluation metric is the 0-1 test accuracy defined in \S\ref{sec:phase3-results}: we parse the model’s final scalar score (or abstention token) and compare it against the physician-preferred $\estimatorNew$ labels, treating abstentions as correct only when $\estimatorNew$ is NA. All decision rules (prompts, reward shape, decoding, and accuracy metric) are held fixed across the two training runs; only the training labels used inside $r(\tau)$ differ.

\paragraph{Hardware and implementation summary.}
Both training runs are conducted on an $8\times$H100 GPU node with two 56-core Intel 8480C CPUs for 9 hours, using the \texttt{verl} framework for distributed RL training \cite{verl}. The experiment run in Fig. \ref{fig:rl-result-main} consumes $2 \times (8\times 9) = 144$ GPU hours; if including the sensitivity analysis, our experiment consumes $144 \times (3 \times 3) = 1296$ GPU hours.

\subsection{Full Training Dynamics on 887-Instance Test Set}\label{app:RL-887}

The main text (\S\ref{results:RL}) reports the RL experiment results on the 50 physician-labeled instances. Here we present the full training dynamics evaluated on the larger 887-instance held-out test set, graded against $\estimatorNew$. Averaging the final ten evaluation steps, the model trained on $\estimatorNew$ achieves $71.4\%$ accuracy versus $62.6\%$ for the model trained on $\estimatorOrig$, an $8.7$ percentage-point gap (Figure~\ref{fig:rl-result-main}). Sensitivity analysis over a $3\times3$ hyperparameter sweep confirms this difference is directionally consistent across all settings.

Notably, while the label discrepancy between $\estimatorOrig$ and $\estimatorNew$ is roughly 27\% (\S\ref{sec:phase2-results}), the downstream RL performance gap is smaller ($8.7\%$), suggesting that the RL process displays some resilience to noisy rewards. This $8.7\%$ swing can also be read as an empirical measure of evaluation uncertainty induced by label choice: changing only the reference labels, while holding the data, model class, and optimization fixed, can move the headline conclusion by $8.7\%$. For context, this difference could eliminate or double the gaps between the top three reported alignment results on MedCalc-Bench (Table~\ref{tab:medcalc_sota} in Appendix~\ref{sec:current-standard}).

\begin{figure}[htbp]
    \centering
    \includegraphics[width=1.0\linewidth]{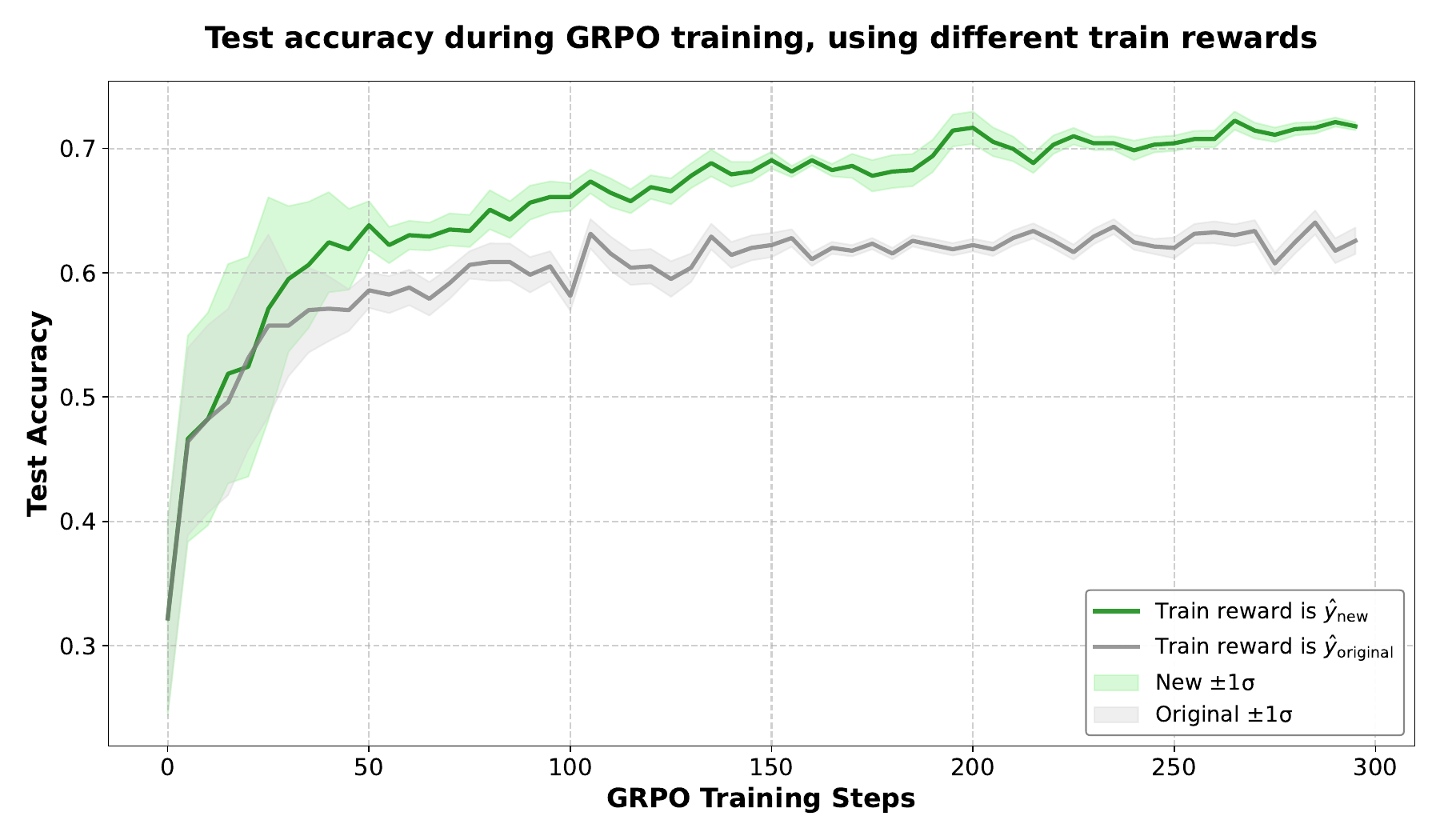}
    \caption{Test accuracy dynamics on the 887-instance held-out set for Qwen3-8B trained via GRPO using the recomputed reference labels (green) versus the original MedCalc-Bench labels (grey). Shaded bands indicate $\pm 1\sigma$ smoothed over a 10-step window. The model trained on $\estimatorNew$ achieves a $+8.7\%$ absolute gain in the final moving averages of test accuracy ($71.4\%$ vs. $62.6\%$).}
    \label{fig:rl-result-main}
\end{figure}

\subsection{Generalization to Additional Medical Evaluation Sets}\label{app:RL-ood}

In this section, we test whether the performance gains reported in \S\ref{results:RL} generalize beyond the MedCalc-Bench test split used in the main experiment. We evaluate both RL-trained checkpoints on three additional medical evaluation sets:
\begin{itemize}
    \item \textbf{MedQA} \cite{medqa}: a medical question-answering benchmark that tests a fundamentally different task (multiple-choice clinical reasoning rather than score computation).
    \item \textbf{MedCalc-v1.2}: a recent revision of MedCalc-Bench posted online by an author of the original benchmark (\href{https://huggingface.co/datasets/ncbi/MedCalc-Bench-v1.2}{HuggingFace link}). Our examination suggests that approximately 70\% of test instances and annotations in the original v1.0 release were replaced, making this a partially overlapping but substantially different version of the same task.
    \item \textbf{MedCalc-Bench-Verified}: another revision (\href{https://github.com/nikhilk7153/MedCalc-Bench-Verified/releases/tag/MedCalc-Bench-Verified}{GitHub link}, December 14 snapshot), also posted by an author of the original benchmark, with similarly extensive changes to the v1.0 release.
\end{itemize}

\paragraph{Regression setup.} We fit a linear probability model with the binary outcome \texttt{Accuracy} (1 if the model's answer is correct, 0 otherwise) as the dependent variable. The treatment variable is the training label source, coded as an indicator for \texttt{orig\_train} (the checkpoint trained on $\estimatorOrig$), with the checkpoint trained on $\estimatorNew$ (\texttt{new\_train}) as the reference category. We also include the base Qwen3-8B checkpoint before either RL run (\texttt{untrained}) as a third level. Control variables are dataset (MedQA, MedCalc-v1.2, MedCalc-Bench-Verified) and decoding configuration (temperature $T$, \texttt{top\_p}, \texttt{top\_k}), which absorb differences in task difficulty and sampling variance across settings.

\paragraph{Results.} The regression was fit on 150{,}048 sampled generations ($(1{,}146 + 990 + 990)$ prompts $\times$ 3 checkpoints $\times$ 2 decoding settings $\times$ 8 sampled trajectories per prompt). The estimated coefficient for \texttt{orig\_train} is $-0.0093$ (95\% CI, $-0.015$ to $-0.004$; $p=0.001$), corresponding to a 0.93 percentage-point accuracy deficit relative to \texttt{new\_train} after controlling for dataset and decoding configuration. Table~\ref{tab:rl-ood-raw-means} reports the unadjusted mean accuracies for every dataset-decoding cell.

\begin{table}[htbp]
\centering
\caption{Mean accuracy (\%) on additional evaluation sets, grouped by dataset, decoding configuration, and checkpoint. The three checkpoints are the model trained on $\estimatorNew$ (\texttt{new\_train}), the model trained on $\estimatorOrig$ (\texttt{orig\_train}), and the base model before RL (\texttt{untrained}). Decoding tuples are reported as ($T$, \texttt{top\_p}, \texttt{top\_k}). The $\Delta$ column reports \texttt{new\_train} minus \texttt{orig\_train} in percentage points.}
\label{tab:rl-ood-raw-means}
\small
\setlength{\tabcolsep}{5pt}
\begin{tabular}{@{}llrrrr@{}}
\toprule
Dataset & ($T$, \texttt{top\_p}, \texttt{top\_k}) & \texttt{new\_train} & \texttt{orig\_train} & \texttt{untrained} & $\Delta$ \\
\midrule
MedQA & $(0.6, 0.95, 20)$ & 79.2 & 78.2 & 75.7 & +1.0 \\
MedQA & $(1.0, 1.0, -1)$ & 78.4 & 78.2 & 74.5 & +0.2 \\
\midrule
MedCalc-v1.2 & $(0.6, 0.95, 20)$ & 64.1 & 63.0 & 30.1 & +1.1 \\
MedCalc-v1.2 & $(1.0, 1.0, -1)$ & 63.1 & 62.9 & 29.2 & +0.2 \\
\midrule
MedCalc-Bench-Verified & $(0.6, 0.95, 20)$ & 64.2 & 62.3 & 29.7 & +1.9 \\
MedCalc-Bench-Verified & $(1.0, 1.0, -1)$ & 63.0 & 61.6 & 29.4 & +1.4 \\
\bottomrule
\end{tabular}
\end{table}

In all six dataset-by-decoding combinations, \texttt{new\_train} outperforms \texttt{orig\_train}, with raw gains ranging from 0.2 to 1.9 percentage points. The effect is directionally consistent but substantially smaller than the $+8.7$ percentage-point gain observed on the in-distribution MedCalc-Bench test split (Figure~\ref{fig:rl-result-main}). This attenuation is expected: the RL fine-tuning signal is most concentrated on MedCalc-style score computation, and a smaller model like Qwen3-8B has limited capacity to transfer fine-tuning gains to tasks beyond its training distribution. Larger models may exhibit stronger generalization of label-quality effects, but testing this hypothesis is beyond the scope of the current study. The key finding is that training on higher-quality labels does not degrade performance on related medical tasks and, if anything, yields a modest, statistically significant improvement.
\section{Two $\nullnotation$ Instances Found by Phase 1 Audit}\label{app:na-examples}

\newcommand{\caseheaderfont}{\normalsize}    
\newcommand{\casetablefont}{\footnotesize}   
\newcommand{\casenotefont}{\scriptsize}      
\newcommand{\casefindingfont}{\footnotesize} 
\newcommand{\caseverdictfont}{\scriptsize}   

In the following pages, we present two MedCalc-Bench test set instances (corresponding to example 1 and 3 in the main text Figure \ref{fig:pipeline1-out2}) whose ground truth would be better labeled as $\nullnotation$ (clinically meaningless or unsafe for an LLM to give a numeric score output), due to two respective causes: \textit{missing data} (required inputs not documented) and \textit{population mismatch} (calculator applied outside validated scope). These instances are drawn from the 279 test instances (26.6\%) that Phase~1 (\S\ref{sec:phase1_audit}) flagged their reference label $\estimatorOrig$ as clinically problematic. The full Phase 1 audit results are available in the \href{https://github.com/junzeye/validate-medcalc-labels/tree/main/data/phase1}{supplementary data release}.

\vspace{16pt}

\noindent
\begin{tikzpicture}
\node[
  draw=headerblue,
  line width=1pt,
  rounded corners=4pt,
  fill=orange!25,
  inner sep=10pt,
  text width=0.99\textwidth,
  align=left
] {%
  {\caseheaderfont\bfseries\textcolor{headerblue}{Case 1\quad$\triangleright$\quad Unanswerable Due to Missing Data}}\\[6pt]
  {\casetablefont\textbf{Patient Note ($C$)}}\\[2pt]
  \colorbox{white}{\begin{minipage}{0.99\textwidth}
  \casenotefont
  ``A 58-year-old male with dyslipidemia, an eight-year history of T2DM, a family history, his mother, of T2DM, with no known micro- or macrovascular complications, was admitted to the emergency department for malaise, epigastric pain, polyuria, and progressive dyspnea which had begun 10 h ago. He had experienced a 2-kg weight loss over the last few days. His usual medications included aspirin 100 mg q24 h, atorvastatin 40 mg q24 h, and metformin 850 mg q8 h, which had been switched to dapagliflozin 20 days before, due to poor glycemic control, with HbA1c 12\% (108 mmol/mol). His vital signs included a heart rate of 122 bpm, respiratory rate 33 rpm, blood pressure 142/70 mmHg, temperature 36.1°C, and body mass index 22.5 kg/m2. On physical examination, somnolence, dry skin and mucous membranes, a Kussmaul breathing pattern, and a capillary refill of 3 sec were observed. Blood tests revealed hemoglobin 17.1 g/dL (13.5–18), leukocytes 19.5 × 103 (4–10 × 103), platelets 296 × 103 (150–450 × 103), glucose 248 mg/dL (60–100), creatinine 0.97 mg/dL (0.67–1.17), sodium 136 mmol/L (135–145), potassium 4.7 mmol/L (3.5–5.5), chloride 101 mmol/L (95–112), phosphate 4.9 mg/dL (2.5–4.5), amylase 70 U/L (10–115), lipase 28 U/L (1–67), pH 6.95 (7.35–7.45), pCO2 23 mmHg (35–45), HCO3 5 mmol/L (22–26), lactate 1.8 mmol/L (0–1.5), urine ketone bodies $>$150 mg/dL (0–0), CK 112 U/L (1–190), CK-MB 7.3 ng/mL (0.1–5), and troponin I 0.07 ng/mL (0.001–0.05). The electrocardiogram (EKG) showed sinus rhythm with right bundle branch block, and nonspecific repolarization abnormalities. Because of the right bundle branch block was not previously known, a new troponin test was performed six hours later with a peak value of 4.28 ng/mL. Treatment with crystalloids, continuous infusion of intravenous insulin, and administration of potassium and sodium bicarbonate were begun in the emergency room (ER). Due to a poor response over the first two hours, with the persistence of lactic acidosis, the patient was transferred to the intensive care unit (ICU), where more aggressive rehydration with crystalloids was started, without further modifications of the original therapeutic plan.
Two days later, the patient was discharged from the ICU to the endocrinology ward. Because of his coronary risk factors and the elevated troponin on admission, a coronary angiography was performed, showing triple-vessel disease. Successful bypass surgery without extracorporeal circulation was performed a few days later, with internal mammary artery grafts to the anterior descendent and marginal obtuse arteries and a saphenous vein graft to the right coronary artery. He was discharged 3 days later on Lantus™ Sanofi-aventis S.p.a Valcanello,03012 Anagni (FR), Italia (insulin glargine) 20 IU and Insulina Novorapid™: Novo Nordisk A/S. Hallas Allé, DK-4400. Kalundborg, Dinamarca (insulin aspart) 6-4-4-0 IU subcutaneous insulin, aspirin, clopidogrel, enalapril, bisoprolol, atorvastatin, and furosemide. The oral antidiabetic treatment with dapagliflozin was not restarted. During follow-up by endocrinology, C-peptide, anti-GAD, and IA-2 antibodies were required due to normal BMI and lack of family history suggested T1DM or a latent autoimmune diabetes of adults (LADA) rather than T2DM as etiology of his diabetes. C-peptide was 1 mg/dL (0.9–7.1) with glycemia 214 mg/dL (60–100), anti-GAD $<$5 U/mL (0–12.5), and anti-IA-2 $<$7.5 U/mL (0–7.5).''
  \end{minipage}}\\[12pt]
  {\casetablefont
  \begin{tabular}{@{}p{5.2cm}p{0.65\textwidth}@{}}
    \textbf{Score Question ($q$):} & Albumin Corrected Anion Gap (in mEq/L)\\[2pt]
    \textbf{Original Reference Label ($\estimatorOrig$):} & -2.75\\[2pt]
    \textbf{Original Label Metadata:} & \{`Chloride': [101.0, `mmol/L'], \colorbox{yellow}{`Albumin': [17.1, `g/dL']}, `Bicarbonate': [5.0, `mmol/L'], `Sodium': [136.0, `mmol/L']\}\\
  \end{tabular}}\\[10pt]
  \textbf{Phase 1 AI Auditor Agent's Analysis (Excerpt)}\\[4pt]
  \small...the GPT incorrectly identified ``hemoglobin 17.1 g/dL'' as the albumin value.** This is a major error. An albumin level of 17.1 g/dL is physiologically impossible......**Conclusion:**\textbackslash{}n    The GPT's calculation is mathematically correct *if* the inputs were valid. However, the key input of albumin is fundamentally wrong. The model mistook the hemoglobin value for the albumin value. Since the patient's true albumin level is not provided in the text, it is impossible to calculate the albumin-corrected anion gap. The final answer of -2.75 is therefore incorrect and clinically meaningless.\textbackslash{}n\textbackslash{}n\textless answer\textgreater\textbackslash{}nNo\textbackslash{}n\textless /answer\textgreater\\[10pt]
};
\end{tikzpicture}

\vspace{20pt}

\noindent
\begin{tikzpicture}
\node[
  draw=headerblue,
  line width=1pt,
  rounded corners=4pt,
  fill=pink!25,
  inner sep=10pt,
  text width=0.99\textwidth,
  align=left
] {%
  {\caseheaderfont\bfseries\textcolor{headerblue}{Case 2\quad$\triangleright$\quad Population Mismatch}}\\[6pt]
  {\casetablefont\textbf{Patient Note ($C$)}}\\[2pt]
  \colorbox{white}{\begin{minipage}{0.99\textwidth}
  \casenotefont
  ``A 55-year-old woman presented with a past medical history of smoking until five years ago (60 pack/year), arterial hypertension, chronic obstructive pulmonary disease with no supplementary oxygen use, and chronic hepatitis B infection with no history suggestive of cirrhosis. Her chief complaint was one year of abdominal discomfort associated with an epigastric mass and weight loss of 30 kg. Simple and contrasted abdominal computed tomography (CT) showed hepatomegaly with focal lesions, the largest being 15 cm in segment VIII, and satellite lesions of up to 5.5 cm in segment VII and 3.5 cm in segment III, with similar morphologic characteristics. Abdominal magnetic resonance imaging (MRI) showed a liver with no signs of chronic disease, and an infiltrative dominant lesion of 11 x 8 cm in segments IV and V, associated with smaller lesions in both lobules of 1.6 cm in segment III, and 0.9 cm in segments VII and VIII (Figure ). Subsequently, the patient presented aggressiveness and disorientation, where hypoglycemia was documented (33 mg/dL) and treated with dextrose 10\% with complete symptom resolution.

Physical examination revealed cachectic facies, and an abdominal mass of approximately 13 cm located in epigastrium and right hypochondrium, associated with hepatomegaly. Initial laboratory were: white blood count of 9900/uL, neutrophils 8000/uL, lymphocytes 1700/uL, monocytes 300/uL, red blood cells 5,540,000/uL, hemoglobin 16.6 g/dL, hematocrit 49.6\%, mean corpuscular volume 89.6 fL, platelet count 357,000/uL, international normalised ratio (INR) 1.05 (normal range (NR) 0.8-1.4), partial thromboplastin time 33 seconds (NR 25-35), serum potassium 3.3 mmol/L, chloride 104 mmol/L (NR 98-107), sodium 141 mmol/L (NR 136-145), total bilirubin 1.01 mg/dL (NR 0.2-0.9), direct bilirubin 0.48 mg/dL (NR 0-0.3), alkaline phosphatase 404 U/L (NR 38-110), gamma-glutamyl transferase (GGT) 1017.6 U/L (NR 10-38), lactic dehydrogenase (LDH) 293 U/L (NR 120-246), alanine transaminase (ALT) 124 U/L (NR 9-52), aspartate transaminase (AST) 226 U/L (NR 14-36), serum creatinine 0.22 mg/dL (NR 0.5-0.8), serum alpha fetoprotein (AFP) 23184.1 ng/mL (NR 0.8-1), carcinoembryonic antigen 0 ng/mL, CA 19-9 antigen 26.86 U/mL (NR 0-37), cortisol 8.9 ug/dL (NR 5.27-22.45), blood ketones 1.0 mg/dL (NR $<$ 3.5), serum glucose 17 mg/dL (NR 74-106), baseline insulin 0.40 uU/mL (NR 2.6-24.9), IGF-I $<$ 15 ng/mL (NR 45-210), C peptide 0.0407 ng/mL (NR 1.1-4.4), IgM antibodies against hepatitis B core antigen non-reactive, antibodies against hepatitis B surface antigen $<$ 3.10 mUI/mL (NR 0-10), antibodies against hepatitis C non-reactive, hepatitis B surface antigen (HBsAg) $>$ 1000 (NR 0-1). Given our patient’s profile, a “big” IGF-II measurement was attempted, but the assay is not available in Colombia''......
  \end{minipage}}\\[12pt]
  {\casetablefont
  \begin{tabular}{@{}p{5.2cm}p{0.65\textwidth}@{}}
    \textbf{Score Question ($q$):} & Sodium Correction for Hyperglycemia (in mEq/L)\\[2pt]
    \textbf{Original Reference Label ($\estimatorOrig$):} & 139.392\\[2pt]
    \textbf{Original Label Metadata:} & \{`Glucose': [33.0, `mg/dL'], `Sodium': [141.0, `mmol/L']\}\\
  \end{tabular}}\\[10pt]
  \textbf{Phase 1 AI Auditor Agent's Analysis (Excerpt)}\\[4pt]
  \small...The patient's glucose level of 33 mg/dL is severe **hypoglycemia** (low blood sugar), not hyperglycemia. Applying a correction for hyperglycemia in a patient who is hypoglycemic is clinically inappropriate and nonsensical. The physiological basis for the correction (osmotic dilution from high glucose) is absent......\#\#\# Conclusion\textbackslash{}n\textbackslash{}nAlthough the GPT performed the mathematical calculation correctly based on its chosen formula, the application of this specific calculator is fundamentally incorrect due to the patient's state of hypoglycemia. A correction for hyperglycemia cannot and should not be applied when the patient's glucose is low. The premise for the calculation is invalid. Therefore, the provided ``Ground Truth Answer'' is clinically incorrect.\textbackslash{}n\textbackslash{}n\textless answer\textgreater\textbackslash{}nNo\textbackslash{}n\textless /answer\textgreater\\[10pt]
  %
};
\end{tikzpicture}

\section{MedCalc-Bench Usage in Anthropic Press Release}\label{app:claude}
In Anthropic's ``Claude for Healthcare'' press release \cite{anthropic_press}, referenced earlier in the main text, the company did not provide the evaluation source code or the version number of MedCalc-Bench dataset with which their models obtained the reported accuracies, i.e., whether it was the original peer-reviewed version published in NeurIPS 2024 (\texttt{v1.0}), or a recent GitHub/HuggingFace revision released by the same authors without a technical report on how their reference labels' curation methodology changed (\texttt{v1.2}).\footnote{Appendix 
\S\ref{app:concur-work} provides more context on the distinction between the two MedCalc-Bench versions.}
Identifying the dataset version used by Anthropic is essential for gauging the clinical validity of their LLMs' benchmark results. To balance inference cost with experiment relevance, we tried to independently evaluate Anthropic's \verb|Haiku 4.5| and \verb|Sonnet 4.5| models using the Claude API, and then compare our results with their results shown in Figure \ref{fig:claude-medcalc}.

\subsection{Reproducing Anthropic's Results on MedCalc-Bench}
\label{sec:anthropic-medcalc}

This experiment has two treatment groups, one evaluating the two said Claude models on the test set of MedCalc-Bench \href{https://web.archive.org/web/20251115234611/https://raw.githubusercontent.com/ncbi-nlp/MedCalc-Bench/5523b7785c004912e76ca4db19efbd7367b23c40/dataset/test_data.csv}{\texttt{v1.0}} (1047 instances), and the other evaluating the two Claude models on the test set of MedCalc-Bench \href{https://web.archive.org/web/20260118051301/https://raw.githubusercontent.com/nikhilk7153/MedCalc-Bench-Verified/cda58b7be721f0e9aab49aec5afd19b814c1d657/datasets/test_data.csv}{\texttt{v1.2}} (1100 instances). We set the inference hyperparameters according to their press release's two footnotes (see bottom of Figure \ref{fig:claude-medcalc}), which says ``\textit{Claude 4.5 models evaluated with extended thinking (64k tokens) and native tool use}'' and ``\textit{(with Python code execution})'' for MedCalc-Bench. Figure \ref{fig:claude-eval-prompt} shows the prompt template we used for all API calls.

\subsection{Our Experiment Results and Analysis}

Figure~\ref{fig:medcalc-reproduction} compares Anthropic's officially reported MedCalc-Bench accuracy with our independent reproduction attempts using both dataset versions. On MedCalc-Bench v1.0, we obtained 50.7\% accuracy for Haiku 4.5 and 62.6\% for Sonnet 4.5 (average of two runs: 62.7\% and 62.5\%). These results are closer to Anthropic's reported values of 47.9\% and 54.4\% than our v1.2 results, which were substantially higher at 65.3\% and 80.5\%, respectively.

The proximity of our v1.0 results to Anthropic's official figures suggests that the company likely evaluated their models on the original NeurIPS 2024 dataset rather than the later GitHub revision. However, a modest discrepancy remains: our v1.0 reproduction exceeds their reported accuracy by 2.8 percentage points for Haiku 4.5 and 8.2 percentage points for Sonnet 4.5. This gap may reflect differences in prompt formatting, answer extraction logic, or other undisclosed evaluation details.

As a sensitivity analysis aside from the experiment treatment, we also tested whether turning on the web search tool, which Anthropic's press release left unmentioned its usage, affected performance. On v1.0, adding web search improved Haiku 4.5 accuracy from 50.7\% to 53.3\% (+2.6pp) but slightly decreased Sonnet 4.5 accuracy from 62.6\% to 61.2\% ($-$1.4pp). On v1.2, turning on web search yielded gains of +4.3pp for Haiku 4.5 (65.3\% to 69.6\%) and +1.6pp for Sonnet 4.5 (80.5\% to 82.1\%). These modest improvements suggest that web search provides supplementary medical knowledge grounding but does not account for the large version-dependent accuracy differences observed.
\newline

\begin{figure}[htbp]
    \centering
    \includegraphics[width=0.9\linewidth]{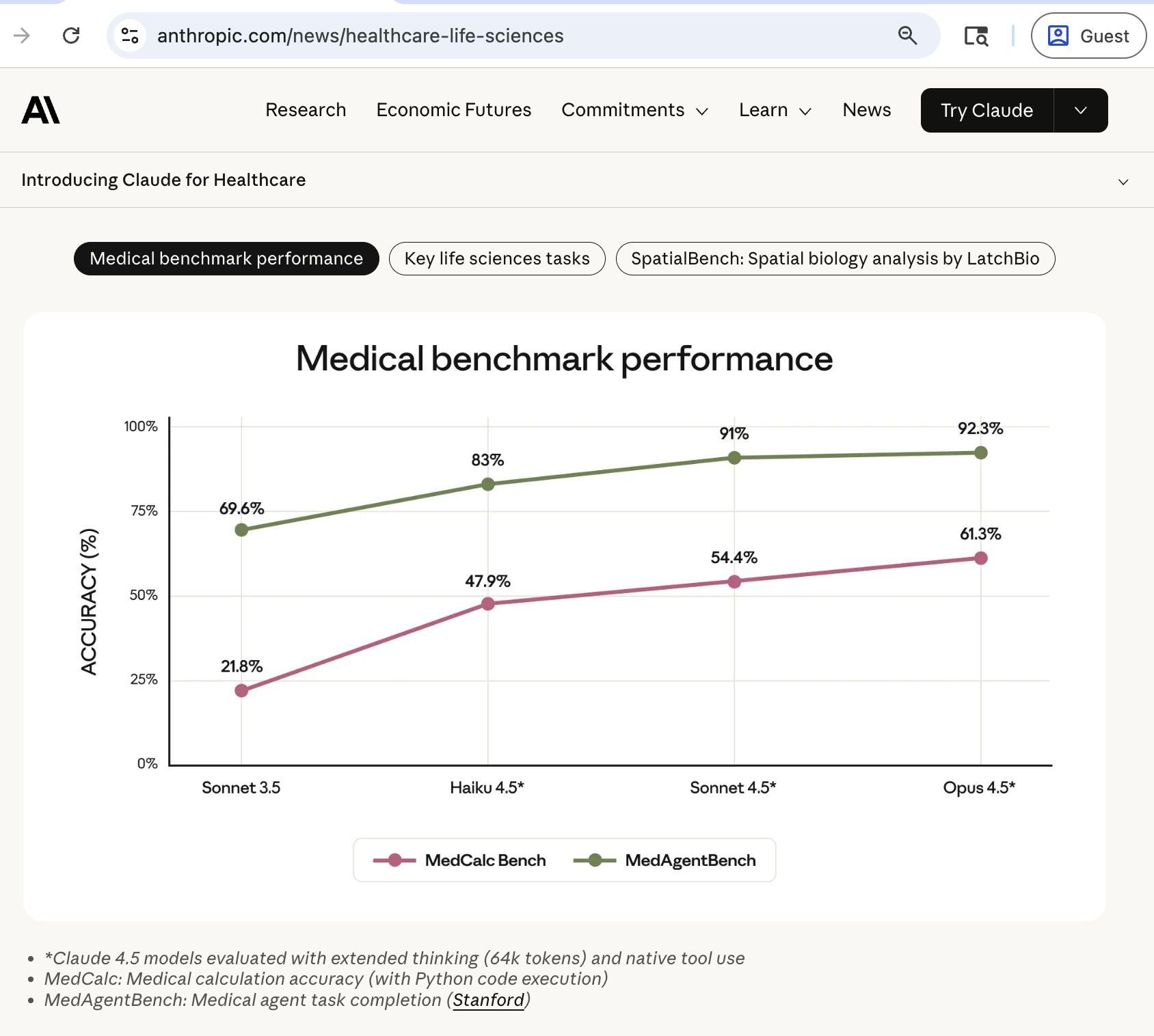}
    \caption{Benchmark performance of four Claude-series models was reported by Anthropic in Exhibit \#1 of their Jan 11, 2026 press release \cite{anthropic_press}. Screenshot taken on Jan 17, 2026. We configured our API call pipeline to match the setup described in the first two bullet-point footnotes. MedAgentBench \cite{medagentbench} is unrelated to our study.}
    \label{fig:claude-medcalc}
\end{figure}

\begin{figure}[htbp]
    \centering
    \includegraphics[width=0.85\textwidth]{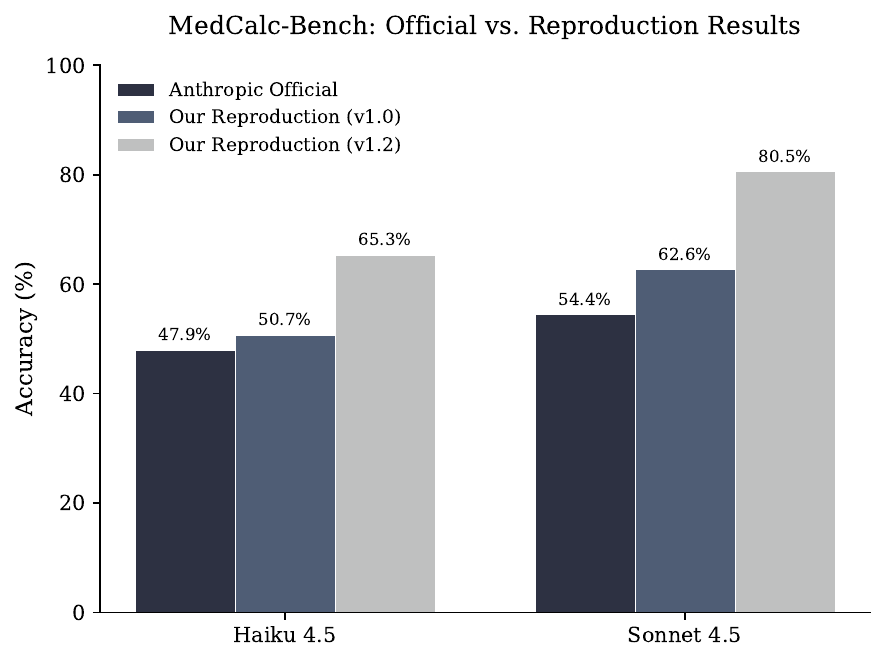}
    \caption{MedCalc-Bench accuracy comparison between Anthropic's official results (shown in Figure \ref{fig:claude-medcalc}) and our independent reproduction using both dataset versions. Our v1.0 reproduction (50.7\% for Haiku 4.5; 62.6\% for Sonnet 4.5) better matches Anthropic's official values (47.9\%; 54.4\%), suggesting they likely evaluated on the original NeurIPS 2024 dataset. Results on v1.2 are substantially higher, indicating that dataset version has a large effect on reported accuracy. All evaluations used extended thinking (64k tokens) and Python code execution, matching the configuration described in Anthropic's press release.}
    \label{fig:medcalc-reproduction}
\end{figure}

\begin{figure}[htbp]
\centering
\begin{tikzpicture}[
  font=\small,
  node distance=6mm,
  stage/.style={draw, rounded corners=2pt, fill=blue!6, align=left, inner sep=3pt, minimum width=38mm},
  arrow/.style={-stealth, thick, draw=black!60},
  every node/.style={align=left}
]

\node[draw, rounded corners=1pt, fill=gray!5, anchor=north west, inner sep=6pt, font=\ttfamily\scriptsize] (bg) at (0,0) {%
\begin{minipage}{0.97\textwidth}
Claude API System Prompt:\\[2pt] 
\colorbox{gray!25}{%
\begin{minipage}{0.99\textwidth}
(LEFT EMPTY)
\end{minipage}%
}\\[4pt]

Claude API User Prompt Template:\\[2pt] 
\colorbox{gray!25}{%
\begin{minipage}{0.99\textwidth}
\ttfamily\scriptsize
You are a medical professional solving a clinical calculation problem.\\

**Patient Note:**\\
\{patient\_note\}\\

**Question:**\\
\{question\}\\

**Instructions:**
\begin{enumerate}
    \item Read the patient note carefully and extract all relevant clinical values.
    \item Identify the appropriate medical formula or calculation method needed.
    \item Use Python code to perform the calculation accurately. Show your work step by step.
    \item After completing your calculation, provide your final numerical answer.
    \item Your final answer MUST be enclosed in <answer></answer> tags. For example, if your calculated answer is 42.5, you would write: <answer>42.5</answer>
\end{enumerate}

Now solve the problem:
\end{minipage}%
}
\end{minipage}%
};

\end{tikzpicture}
\caption{API call prompt template, applied identically to both treatment groups in \S\ref{sec:anthropic-medcalc}.}
\label{fig:claude-eval-prompt}
\end{figure}

\end{appendices}

\end{document}